\documentclass[journal]{IEEEtran}
\usepackage{hyperref}
\usepackage{cite}
\usepackage{amsmath,amssymb,amsfonts}
\usepackage{algorithm}
\usepackage{algpseudocode}
\usepackage{graphicx}
\usepackage{caption}
\usepackage{subcaption}
\usepackage[compact]{titlesec}  
\titlespacing{\section}{2pt}{2pt}{2pt}
\usepackage{textcomp}
\usepackage{xcolor}
\usepackage{comment}
\usepackage{hyperref}
\def\BibTeX{{\rm B\kern-.05em{\sc i\kern-.025em b}\kern-.08em
    T\kern-.1667em\lower.7ex\hbox{E}\kern-.125emX}}
\usepackage{bm}
\usepackage{enumitem}
\usepackage[normalem]{ulem}
\usepackage{cases}
\ifCLASSINFOpdf

\else

\fi

\usepackage{multirow}
\usepackage{array}
\usepackage{adjustbox}
\usepackage{lipsum}
\usepackage{makecell}

\begin{document}

\title{Digital Twin in Industries:  A Comprehensive Survey}

\author{Md Bokhtiar Al Zami,
        Shaba Shaon,
        Vu Khanh Quy,
        Dinh C. Nguyen
\thanks{

Md Bokhtiar Al Zami, Shaba Shaon, and Dinh C. Nguyen are with ECE Department,  University of Alabama in Huntsville, Huntsville, AL 35899, USA, emails: (mz0024@uah.edu, ss0670@uah.edu, dinh.nguyen@uah.edu).}
\thanks{Vu Khanh Quy is with the Faculty of Information Technology, Hung Yen University of Technology and Education, Hung Yen 160000, Vietnam, email: (quyvk@utehy.edu.vn).}
}

\markboth{}%
{}

\maketitle


\begin{abstract}

Industrial networks are undergoing rapid transformation driven by the convergence of emerging technologies that are revolutionizing conventional workflows, enhancing operational efficiency, and fundamentally redefining the industrial landscape across diverse sectors. Amidst this revolution, Digital Twin (DT) emerges as a transformative innovation that seamlessly integrates real-world systems with their virtual counterparts, bridging the physical and digital realms. In this article, we present a comprehensive survey of the emerging DT-enabled services and applications across industries, beginning with an overview of DT fundamentals and its components to a discussion of key enabling technologies for DT. Different from literature works, we investigate and analyze the capabilities of DT across a wide range of industrial services, including data sharing, data offloading, integrated sensing and communication, content caching, resource allocation, wireless networking, and metaverse.  In particular, we present an in-depth technical discussion of the roles of DT in industrial applications across various domains, including manufacturing, healthcare, transportation, energy, agriculture, space, oil and gas, as well as robotics. Throughout the technical analysis, we delve into  real-time data communications between physical and virtual platforms to enable industrial DT networking. Subsequently, we extensively explore and analyze a wide range of major privacy and security issues in DT-based industry. Taxonomy tables and the key research findings from the survey are also given, emphasizing important insights into the significance of DT in industries. Finally, we point out future research directions to spur further research in this promising area.

\end{abstract}


\IEEEpeerreviewmaketitle

\section{Introduction}
The Industrial Revolution marked the beginning of a new age of technological innovation and automation empowered by recent advances in Industrial Internet of Things (IIoT) \cite{10239369}. This pivotal era laid the groundwork for the advanced industrial processes that continue to evolve in today's modern economy. In recent years, digital twin (DT) has emerged as a key enabler of this evolution, allowing industries to bridge the physical and digital realms through bidirectional communications, real-time simulations and monitoring. By optimizing operations and enhancing decision-making, DT is driving smarter manufacturing, predictive maintenance, and more efficient infrastructure management \cite{9711524}.

Recently, DT models have captured significant attention for their robust potential and versatility, offering substantial benefits across a wide array of sectors, including healthcare, education, agriculture, and manufacturing \cite{9882337},\cite{10345669}. Their ability to provide real-time insights, optimize processes, and enhance decision-making has driven their adoption and exploration in numerous additional fields \cite{9359733}. With their innovative operational approach, DT models offer various important benefits for industrial applications
that fall under the scope of distinct deployment levels as follows, each reflecting a different degree of virtualization \cite{10106261}:

\begin{itemize}
    \item \textit{Monitoring:} At this level, DT provides a virtual representation of a physical object. This function offers the ability to monitor the operation of physical entities through controlling its digital counterparts in the digital platform.
    \item \textit{Simulation:} DT serves as a simulator for physical objects, enabling understanding, prediction, and optimization. The virtual model adapts to changes, though these changes do not impact the physical object.
    \item \textit{Operation:} 
    This level features bidirectional communications between physical objects and their DTs via Ethernet, Wi-Fi, or wireless cellular networks, with state changes reflected in both the virtual and physical entities.
\end{itemize}

\begin{figure*}[ht!]
     \centering
     \includegraphics[width=0.99\textwidth]{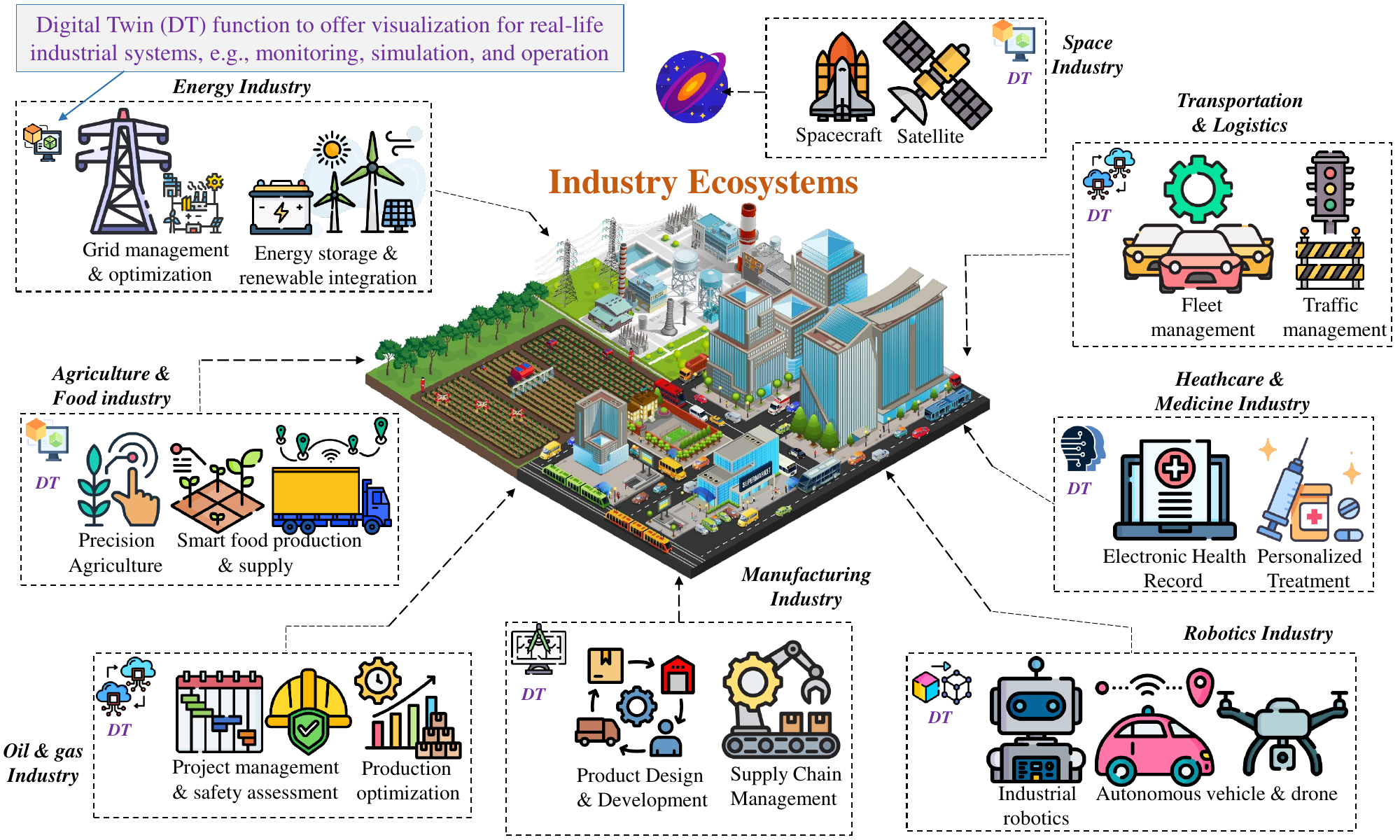} 
     \caption{The convergence of   DT and industries. }
     \label{overview_fig}   
\end{figure*}

Leveraging its distinctive benefits, DT technology has been suggested for a wide range of industrial applications, including smart manufacturing, smart healthcare, smart transportation, energy management, satellite communication, etc. For example, DT plays a crucial role in implementing smart manufacturing by creating digital replicas of manufacturing systems, machines, and processes across industries \cite{liu2021digital}. In healthcare, DT enhances patient data management and personalizes treatment plans, while also improving surgical planning \cite{haleem2023exploring}. In transportation and logistics, DTs leverage IIoT networks and wireless communication to improve the efficiency of monitoring and optimizing resources\cite{martinez2021digital}. DT s impact in agriculture and food production is supported by wireless sensor networks that enable precise monitoring and predictive analysis \cite{sleiti2022digital}. DT has also proved its potential in agriculture and food production by supporting precise monitoring and predictive analysis, which boosts productivity and sustainability \cite{ghandar2021decision}. In satellite operations, DT improves accuracy and reliability through advanced monitoring and predictive maintenance of assembly processes and network performance \cite{leutert2024ai}. Furthermore, in the management of autonomous vehicles, drones, and smart ports, DT utilizes cellular networks and wireless communication to enhance navigation, safety, and operational efficiency \cite{denk2022generating}. All these notable progress and achievements in DT applications across diverse industries highlight the ideal moment to delve deeper into this revolutionary field of research. The overview of the integration of DT across various industries, which will be presented in this paper, is illustrated in Fig. ~\ref{overview_fig}. DT with its great technical potential has significantly transformed many industrial sectors, from energy, transportation to manufacturing and robotics.

\subsection{Comparison and Our Contributions}
Driven by the latest developments in DT technology and its integration with various applications, recent surveys have been introduced to explore the transformative impact and emerging trends in this field. For instance, the work in \cite{fuller2020digital} contributes to the field of DT concept, emphasizing its integration with Industry 4.0 and key enabling technologies. It highlights the rapid evolution of DT applications, particularly within the manufacturing sector, and the role of key enabling technologies such as AI and the IIoT. Similarly, the authors in \cite{barricelli2019survey} analyzed current definitions and core characteristics of DT technology, exploring its application across various domains. They also presents design implications related to socio-technical aspects and the DT life cycle. Other articles in \cite{minerva2020digital}-\cite{xu2023survey} studied the impact of DT in the context of IIoT and related fields. The researchers in \cite{minerva2020digital} presented an overview of the definitions and features of DT, extending its application in IoTs.  Various application scenarios have also been presented along with software architecture models, and potential evolution paths for DTs, highlighting their role in the broader softwarization process. DT technology within the IIoT was explored in \cite{xu2023survey}, focusing on enabling technologies such as AI and blockchain. The study discussed real-world applications, architectures, and models, investigating advanced technologies for intelligent and secure DT-IIoT, and proposing software tools for high-fidelity modeling. Meanwhile, the authors in \cite{wang2023survey} offered a review of  security and privacy issues in generic DT systems along with a discussion of defense approaches. Another paper in \cite{alcaraz2022digital} also concentrated on the security landscape of DT technology within the Industry 4.0 paradigm, focusing on the confluence of various enabling technologies such as cyber-physical systems, IIoT, edge computing, and AI. It classified potential security threats across the functionality layers of DTs, addressing operational requirements for a more thorough understanding of these risks. The paper also offered preliminary security recommendations and approaches to enhance the trustworthy and secure deployment of DTs in industrial settings. The integration of DT in wireless networks was explored in \cite{khan2022digital}, providing an overview of DT technology for wireless systems. It discussed key concepts and taxonomy, and explored design aspects, deployment trends, security, and air interface considerations. Similarly, the work in \cite{zeb2022industrial} contributed by examining the role of DT technology in smart industries, specifically from a communication and computing perspective. It reviewed recent research trends in next-generation wireless technologies (such as 5G and beyond) and computational paradigms like edge and cloud computing. The role of DT technology in 6G communication systems was surveyed in \cite{kuruvatti2022empowering}, critically analyzing deployment in 6G systems, and highlighting potential use cases and applications. The integration and advancement of DT in industries was studied in \cite{tao2018digital}, \cite{melesse2021digital}, and \cite{melesse2021digital}. The authors in \cite{tao2018digital} provided a review of DT applications in the industry, examining the key components and development, along with successful applications in areas such as product design, production, and health management. The work in \cite{melesse2021digital} by conducted a review of DT applications in industrial operations, focusing on production, predictive maintenance, and after-sales services. Another paper in \cite{su2023digital} offered a survey of DT applications in the construction industry, particularly focusing on life cycle management. The comparison of the related works and our paper is summarized in Table ~\ref{tab:taxonomy1}.

\begin{table*}
    \centering
    \caption{Existing surveys on DT-related topics and our new contributions.}
    \label{tab:taxonomy1}
    \small
    \begin{adjustbox}{max width=\textwidth}
    \begin{tabular}{|m{0.5cm}|m{2.2cm}|m{8.2cm}|m{6.3cm}|}
        \hline
        \textbf{Ref.} & \textbf{Topic} & \textbf{Key contributions} & \textbf{Limitations} \\ \hline
        \cite{fuller2020digital} & DT Concept & A review of DT technology, focusing on its development, key models, and enabling technologies. & This paper does not present detailed solutions for addressing complex engineering challenges in DT integration, \\ \hline
        
        \cite{barricelli2019survey} & DT Concept & An analysis of DT definitions, key characteristics, and applications across domains, with design insights. & This article lacks detailed application services and solutions to DT challenges. \\ \hline

     \cite{minerva2020digital} & DT for IoT & A survey of DT-driven IoT, highlighting AI, blockchain, and high-fidelity modeling. & The paper lacks practical details and guidelines for integrating DT-IIoT with emerging technologies. \\ \hline
     
        \cite{xu2023survey} & DT for IIoT & A review of DT-IIoT, discussing its enabling technologies such as AI and blockchain, architectures, and real-world applications & The paper lacks in-depth exploration of practical implementation challenges and scalability in real-world industrial settings.  \\ \hline

        \cite{wang2023survey} & DT and security & A review on DT, categorizing security threats and highlighting key challenges. & The survey lacks practical implementation strategies and detailed assessments of real-world security challenges. \\ \hline

        \cite{alcaraz2022digital} & DT and security & A review on classifying potential security threats across different functionality layers of DT. & The paper provides only a preliminary exploration of security solutions, lacking specific case studies to validate the recommendations. \\ \hline

        \cite{khan2022digital} & DT for wireless & An overview of DTs in wireless systems, including concepts, design, and taxonomy. & Practical implementation details and solutions for DTs in wireless systems are not thoroughly explored. \\ \hline

        \cite{zeb2022industrial} & DT for wireless & An examination of DT’s role in industries, including integration with next-gen wireless tech and computational intelligence. & The article does not provide specific case studies and detailed methods for DT implementation and practical deployment challenges. \\ \hline
        
        \cite{kuruvatti2022empowering} & DT for wireless & An exploration of DT's role in 6G, reviewing literature, deployment, use cases, standards activities, and research challenges. & The paper does not provide methods for DT integration into 6G and detailed strategies for research challenges. \\ \hline

        \cite{tao2018digital2} & DT for Industry & A review on the development, key components, and applications of DTs in industry. & The paper does not provide detailed exploration of how to overcome practical challenges in implementing DTs across diverse industrial environments. \\ \hline
        
        \cite{melesse2021digital} & DT for Industry & A survey on DT in industrial operations, emphasizing its role in production and predictive maintenance. & The paper lacks a comprehensive survey on industrial DT applications, e.g., agriculture, space, and robotics. \\ \hline
        
        \cite{su2023digital} & DT for Industry & A review of DT in the construction industry, highlighting its potential and challenges across different construction phases. & The paper paper only focuses on one kind of industry, and hence is not comprehensive. \\ \hline
        
        \textit{Our work} & DT in industry & A comprehensive survey on DT in industries. Specifically, \begin{itemize}
        \item We present the fundamentals, components, and discuss enabling technologies of DT. 
        \item We carry out an in-depth analysis of DT services in industries, including industrial data sharing, data offloading, integrated sensing and communication, content caching, resource allocation, wireless networking, and the metaverse. 
        \item We present a holistic discussion of the roles of DT in industrial networks across important domains, e.g., healthcare, manufacturing, oil and gas, transportation, energy, agriculture, space, and robotics, where the role of communication techniques is highlighted for enabling industrial DT networks.
        \item We  investigate and discuss a wide range of major privacy and security issues in DT-based industry.
        \item Taxonomy tables and main findings are given to provide insights into driving DTs for industries. Challenges and future research directions are also highlighted.
        \end{itemize} &  \\ \hline
    \end{tabular}
    \end{adjustbox}
\end{table*}

Despite such research efforts, \textit{they lack a comprehensive and dedicated survey of DT for industrial services and applications.} In particular, the potential of DT in industrial services, such as data sharing, data sensing and offloading, content caching, resource management,  wireless networking, metaverse, etc., has been under-explored in the open literature \cite{tao2018digital}-\cite{melesse2021digital}. Moreover, a holistic discussion of DT applications across diverse industries is still missing in \cite{minerva2020digital}-\cite{xu2023survey}. \textit{It is noting that such existing works provide only a partial analysis of DT-based industrial applications}, while a comprehensive survey for all important application domains, ranging from robotics, and manufacturing to agriculture and space, has not been investigated.

Motivated by these limitations, this paper presents a more comprehensive survey of the integration of DT in industrial networks, including both industrial services and applications. We emphasize a thorough discussion on bidirectional communication between physical entities and their digital counterparts within a unified DT platform for each industrial use case, providing valuable insights into industrial DT networking operations. Security and privacy issues in the DT-based industry are also highlighted. These are also our key novelties that make our paper fundamentally different from all related literature works. To this end, the main contributions of this article are outlined as follows:

\begin{enumerate}

    \item We provide a comprehensive survey on the use of DT in industries where its key fundamentals, components, and main enabling technologies are discussed. 
    
    \item We present a detailed discussion on the roles of DT in key industrial services, namely data sharing, data offloading, integrated sensing and communication, content caching, resource allocation, wireless networking, as well as emerging concepts like metaverse, with a particular focus on the nature of communication and networking protocols among systems, machines, processes and their digital counterparts.
    \item We conduct a holistic investigation of the applications of DT in a wide range of industrial domains, including manufacturing industry, healthcare and medicine industry, transportation and logistics, energy industry, agriculture and food industry, space industry, oil and gas industry, and robotics industry, highlighting the pivotal role of communication and networking technologies in enhancing DT effectiveness. Additionally, we provide taxonomy tables that summarize the key technical aspects, contributions, and limitations of each DT approach employed in industry.
    \item We also explore security and privacy challenges across multiple levels of these industries, including physical, digital, communication, and HMI layers, and proposes countermeasures to address these vulnerabilities effectively. Furthermore, we offer countermeasures aimed at addressing and mitigating these vulnerabilities efficiently.
    \item Drawing from the extensive survey, we highlight research findings and taxonomy tables are also given. Finally, we point out potential future research directions to spur further research in this promising area.
\end{enumerate}

\subsection{Structure of the Survey}
\begin{figure}[!h]
    \centering
    \includegraphics[width = 0.5\textwidth]{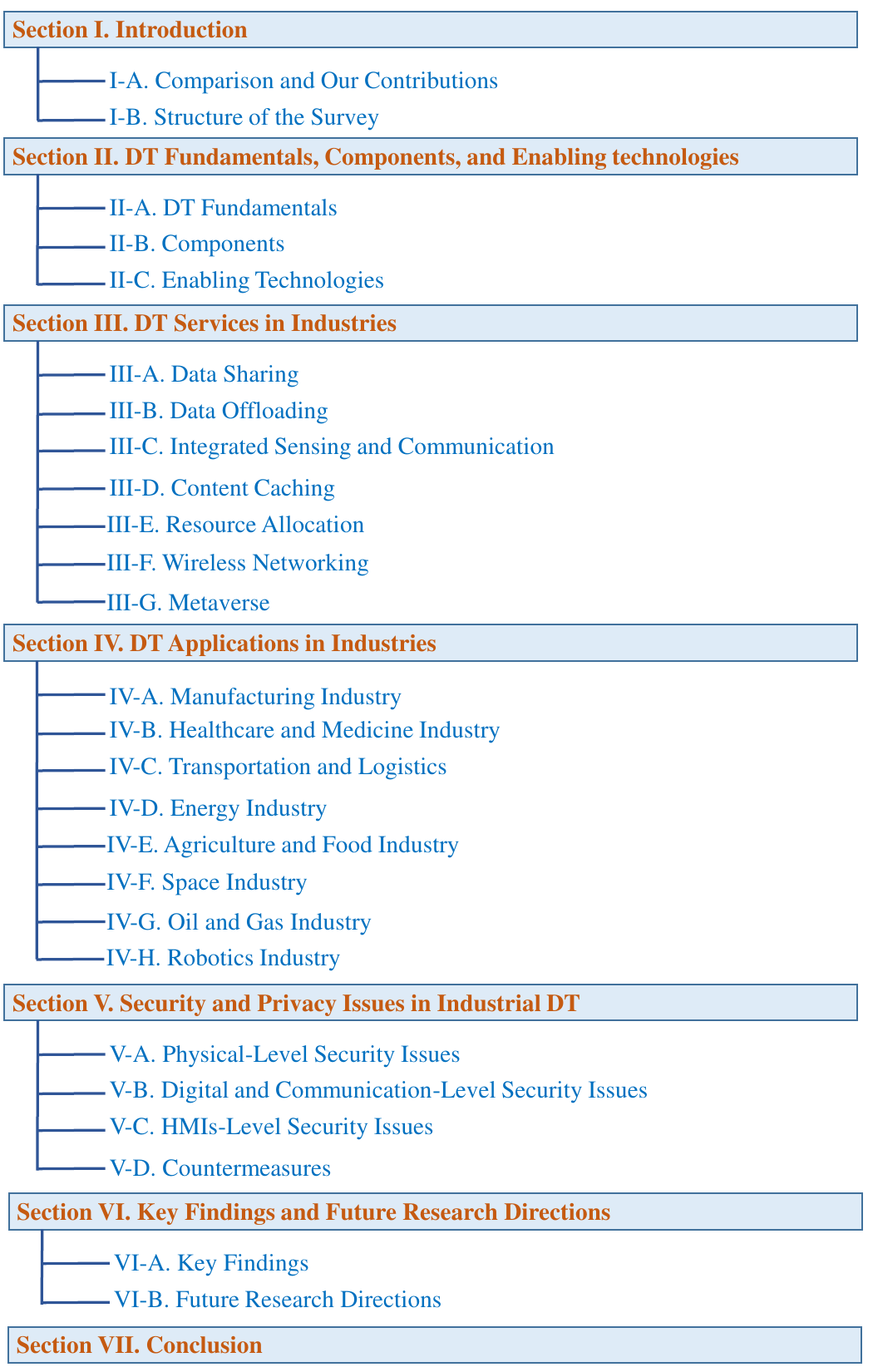}
    \caption{Organization of this article.}
    \label{Organization} \vspace{-10pt}
\end{figure}

This structure of our survey is illustrated in Fig.~\ref{Organization}.  Section \ref{SectionII} reviews the fundamentals of DT, its components, and enabling technologies.  Section \ref{SectionIII} provides an in-depth analysis of DT services across diverse industrial settings. Section \ref{SectionIV} explores the potential of DT across various industrial applications. Section \ref{SectionV} delves into the security issues at various levels within industrial networks.  Section \ref{SectionVI} outlines the key findings and potential directions for future research. Finally, Section \ref{sec:conclude} concludes the article. 


\section{DT Fundamentals, Components, and Enabling Technologies} \label{SectionII}
In this section, we highlight the fundamentals and key components of DT, followed by a discussion of several important enabling technologies of DT. 

\subsection{DT Fundamentals}
DT technology stands at the forefront of the fourth industrial revolution, offering a bridge between the physical and digital worlds. The core concept of a DT involves creating a highly detailed and dynamic digital replica of a physical object, process, or system. This digital counterpart is continuously updated with real-time data from its physical counterpart, enabling a synchronized relationship that enhances understanding, prediction, and control of the physical entity. The foundation of DT technology is built upon several key principles:
\begin{itemize}
    \item \textit{Real-Time Data Integration:} The essence of a DT lies in its ability to integrate real-time data from various sensors and data sources attached to the physical entity. This continuous data flow ensures that the digital model is an accurate and up-to-date representation of its physical counterpart, allowing for real-time monitoring and analysis.
    \item \textit{High-Fidelity Modeling:} Creating an effective DT requires sophisticated modeling techniques that accurately capture the physical and operational characteristics of the real-world entity. These models can range from simple geometric representations to complex simulations that encompass physical behaviors, operational conditions, and even environmental interactions.
    \item \textit{Simulation and Analysis:} DTs leverage advanced simulation capabilities to analyze current conditions and predict future states. By simulating various scenarios and operational changes, they enable stakeholders to test hypotheses, evaluate potential outcomes, and make informed decisions without risking the actual physical entity.
    \item \textit{Interoperability and Integration:} For a DT to be effective, it must seamlessly integrate with other digital systems and platforms. This interoperability ensures that data can be aggregated from multiple sources and that the insights generated by the DT can be disseminated across the organization.
    \item \textit{Lifecycle Management:} DTs are valuable throughout the entire lifecycle of a physical entity, from design and manufacturing through operation and maintenance to decommissioning. By providing insights at each stage, they enable continuous improvement and optimization, enhancing performance and extending the lifecycle of the physical entity.
    \item \textit{Predictive and Prescriptive Capabilities:} Beyond mere representation, DTs utilize predictive analytics to forecast future conditions and prescriptive analytics to recommend optimal actions. These capabilities are driven by machine learning algorithms and AI, which analyze historical and real-time data to identify patterns, predict failures, and suggest preventive measures.
\end{itemize}

DT technology is revolutionizing industries by providing a comprehensive digital perspective on physical assets. This approach not only enhances operational efficiency and reduces downtime but also fosters innovation by enabling the exploration of new operational strategies and business models. As industries continue to evolve, the role of DTs is poised to expand, driving further advancements in how we design, operate, and manage complex systems.

\subsection{Components}
DT technology comprises several critical components that work together to create a functional and effective digital representation of physical entities. These components include the physical entity, the virtual model, the data, the connection, and the services, each playing a vital role in the overall system.

\begin{itemize}
    \item \textit{Physical Entity:} The physical entity is the real-world object, process, or system that the DT represents. It can range from a single piece of machinery to an entire manufacturing plant, a vehicle, or even a human body. The physical entity is equipped with various sensors and data acquisition devices that continuously capture operational and environmental data.
    \item \textit{Virtual Model:} The virtual model is the digital replica of the physical entity. This model can be a geometric representation, a physics-based simulation, or a combination of various modeling techniques. The virtual model is designed to mimic the behavior, performance, and interactions of the physical entity accurately. It serves as the core of the DT, enabling simulations, analyses, and visualizations.
    \item \textit{Data:} Data is the lifeblood of the DT. It includes all the information collected from the physical entity, such as sensor data, operational logs, environmental conditions, and historical performance data. This data is used to update and refine the virtual model continuously, ensuring that it remains an accurate and up-to-date representation of the physical entity.
    \item \textit{Connection:} The connection component refers to the communication infrastructure that links the physical entity with the virtual model. This connection can be established through various means, such as wired or wireless networks, cloud platforms, or edge computing systems. The connection ensures real-time data flow between the physical and digital realms, enabling continuous synchronization and interaction.
    \item \textit{Services:} Services encompass the various applications and tools that leverage the DT for specific purposes. These services include real-time monitoring, predictive maintenance, performance optimization, and decision support. Advanced analytics, machine learning algorithms, and AI-driven insights are often part of these services, providing valuable information to stakeholders for informed decision-making and proactive management.
\end{itemize}

Each component plays a crucial role in the successful implementation and operation of DTs. By seamlessly integrating these components, DT technology provides a powerful framework for understanding, predicting, and optimizing the performance of physical entities in a wide range of industries.

\subsection{Enabling Technologies}
While there are many enabling technologies discussed in literature survey papers, here we present three major enabling technologies that directly drive low-latency monitoring (i.e., edge computing), simulation (i.e., machine learning), and operation (i.e., wireless communications). The roles of such important technologies for DT are highlighted below.
\subsubsection{Edge Computing}
Edge computing is a transformative technology that significantly enhances the functionality and efficiency of DT systems. By processing data closer to its source, edge computing reduces latency, improves real-time data handling, and optimizes bandwidth usage, making it an indispensable component of modern DT architectures.
\begin{itemize}
    \item \textit{Reducing Latency:}
    One of the primary advantages of edge computing in DT applications is the dramatic reduction in latency. Traditional cloud computing models often involve transmitting large volumes of data from the physical entity to a centralized server for processing and then back again. This round-trip can introduce significant delays, which are unacceptable in scenarios requiring real-time responsiveness. Edge computing mitigates this issue by performing data processing tasks at or near the source of data generation, enabling instantaneous analysis and action. This is particularly crucial in industrial automation, healthcare monitoring, and other time-sensitive applications \cite{shi2016edge}.
    \item \textit{Enhancing Real-Time Processing:}
   DTs rely on continuous streams of data from sensors and other data sources to maintain an accurate and up-to-date digital representation of the physical entity. Edge computing facilitates this by allowing for the immediate processing of data as it is generated. Edge devices can perform initial filtering, aggregation, and analysis of data, ensuring that only relevant information is sent to central systems for further processing. This not only enhances the speed and efficiency of data handling but also reduces the computational load on central servers \cite{zhang2023digital}.
   \item\textit{Optimizing Bandwidth Usage:}
   In DT environments with numerous sensors and devices generating vast amounts of data, transmitting all raw data to the cloud can be inefficient and costly. Edge computing helps optimize bandwidth usage within DT systems by processing and compressing data locally before sending it over the network. This localized data processing reduces the amount of data that needs to be transmitted, conserving bandwidth and lowering transmission costs \cite{liu2021digital}.

   \item \textit{Enabling Scalability and Flexibility:}
   As DT systems expand day by day, the ability to scale and adapt becomes crucial. Edge computing plays a key role by offering a flexible infrastructure that can grow alongside increasing numbers of physical entities and data volumes. Instead of requiring major overhauls to the IT infrastructure, additional edge devices can be deployed as needed to manage increased computational demands. This modular approach allows DT systems to evolve and adjust in response to changing requirements, to maintain operational efficiency and performance without significant disruption \cite{qi2018modeling}.

\end{itemize}

\subsubsection{Machine Learning (ML)}

The synergy between  ML and DT technology is reshaping how we interact with and optimize complex systems. By embedding ML algorithms into DT frameworks, we can significantly enhance the capabilities of these digital models, enabling more accurate simulations, predictions, and decision-making processes. ML empowers DTs by transforming the static nature of traditional digital models into dynamic, adaptive entities. Through continuous learning from real-time data, ML algorithms allow DTs to evolve and refine their simulations. This ongoing adaptation helps the digital model more accurately reflect the current state and behavior of its physical counterpart, improving the overall precision of the virtual representation \cite{chakraborty2021machine}. One of the primary benefits of integrating ML with DTs is the enhancement of predictive analytics. ML algorithms can process and analyze vast amounts of data to uncover trends and patterns that traditional methods might miss. For example, by utilizing supervised learning techniques, DTs can be trained to predict potential failures or inefficiencies in a system before they occur. This predictive capability is crucial for implementing preventive measures and optimizing system performance, thereby reducing operational disruptions \cite{al2023big}. Unsupervised learning methods also contribute to the efficacy of DTs by identifying previously unknown relationships within the data. These methods help in discovering hidden anomalies or correlations, which can refine the accuracy of simulations and improve the understanding of system dynamics. Such insights are invaluable for adjusting the behavior of DTs in response to emerging data trends \cite{xu2019digital}. Furthermore, the application of reinforcement learning (RL) within DT environments supports real-time decision-making and autonomous optimization. RL algorithms can experiment with various strategies, learn from the outcomes, and adjust operational parameters to enhance performance. This capability is particularly beneficial in environments where timely decisions are critical for maintaining efficiency and achieving optimal results \cite{muller2020dynamic}. The collaboration of ML with DT also enhances the ability to simulate and optimize complex scenarios. By leveraging ML-driven insights, DTs can conduct sophisticated simulations that account for a wide range of variables and interactions. This allows for more comprehensive testing and validation of different operational strategies and design configurations before implementing them in the real world \cite{tao2022digital}.
\subsubsection{Wireless Communications}
The rapid evolution of wireless technologies is crucial in advancing DT systems, particularly in their application across emerging fields such as extended reality, brain-computer interaction, and advanced healthcare. These next-generation applications impose demanding performance requirements, including stringent quality of experience, low latency, and high reliability. Traditional wireless systems often fall short of meeting these diverse needs, creating a compelling case for the integration of DT technology \cite{rathore2021role}. DTs represent a sophisticated amalgamation of virtual models and real-world systems, with wireless communications serving as a fundamental enabler. The advent of advanced wireless technologies, such as 5G and the forthcoming 6G, provides the necessary infrastructure to support the complex and data-intensive requirements of DTs. These technologies offer high-speed connectivity, reduced latency, and improved reliability, which are essential for the seamless operation of DTs. For instance, the ultra-low latency of 5G networks is particularly beneficial for applications requiring near-instantaneous data transmission, such as real-time monitoring and remote control of digital replicas \cite{lin20236g}. The interplay between DTs and wireless communications extends beyond mere data transfer. Edge computing, which involves processing data closer to its source, complements wireless technologies by mitigating latency and alleviating network congestion. This synergy is crucial for applications that demand real-time responses, such as augmented reality experiences and real-time medical diagnostics. By performing preliminary data processing at the edge, these systems ensure that only the most pertinent information is transmitted over the network, thus enhancing overall system efficiency and responsiveness \cite{dong2019deep}. Besides, the integration of ML and AI with DTs, facilitated by advanced wireless networks, provides significant advantages. ML algorithms analyze the extensive data generated by DTs to offer predictive insights, optimize performance, and identify potential issues before they escalate. This integration enables more intelligent and adaptive systems, capable of learning from both historical and real-time data. Wireless technologies play a pivotal role in this process by providing the necessary bandwidth and speed for processing and transmitting large volumes of data \cite{kaur2020convergence}. Security and privacy are critical aspects of DT systems, especially when integrated with wireless communications. The use of blockchain technology for secure data transactions and integrity verification is becoming increasingly relevant. Blockchain provides a decentralized and immutable ledger, ensuring that data exchanged between physical and digital entities remains secure and tamper-proof \cite{nguyen2021federated1}. This is particularly important in applications where data integrity and security are paramount, such as in healthcare or sensitive industrial operations \cite{putz2021ethertwin}.
Designing DT systems involves addressing several technical and logistical challenges. Key considerations include the creation of effective digital models and interfaces that interact seamlessly with physical systems. This involves addressing issues related to the isolation of DTs, ensuring that they operate independently yet cohesively within the broader system framework. Additionally, the deployment of DT systems requires innovative solutions for prototyping, incentive mechanisms, and system decoupling, which are essential for effective integration and scalability \cite{khan2022digital}. Wireless communications also facilitate the expansion and scalability of DT systems. As the number of connected devices and data sources increases, wireless networks can be scaled to accommodate these demands without extensive infrastructure changes. This scalability is crucial for supporting the growing complexity of DT applications and ensuring that they remain effective and responsive as they evolve \cite{he2022security}.

\section{DT Services in Industries} \label{SectionIII}


In this section, we present an in-depth discussion on the roles of DTs in various industrial services, including data sharing, data offloading, integrated sensing and communication, content caching, resource allocation, wireless networking, and metaverse. 

Moreover, we summarize the taxonomy of DT services in industries discussed in this work in Table ~\ref{tab:taxonomy_services} continued to Table ~\ref{tab:taxonomy_services_continued}. 
\subsection{Data Sharing}
Data sharing is a fundamental service in industrial applications, crucial for the seamless transfer of information over shared networks to support various end-user needs. Traditional methods involve transmitting raw industrial data directly, which can be cumbersome and inefficient, particularly when dealing with vast quantities of industrial data generated in real time. This approach often leads to challenges such as high latency, potential data breaches, and difficulties in deriving actionable insights from the data alone. DT technology revolutionizes industrial data sharing by shifting the focus from raw data to actionable insights derived from real-time simulations and analyses \cite{li2023breaking}. Unlike conventional industrial data-sharing methods, DTs create virtual replicas of physical assets, IIoT devices, machines, actuators, processes, or systems. These replicas are continuously updated with real-time industrial data, allowing for the simulation and analysis of various scenarios without impacting the actual physical systems. This means that rather than sharing raw data, industries can share insights and predictive analytics that are derived from these virtual models. The integration of DT technology into industrial data sharing provides several key benefits. First, it significantly reduces latency by enabling real-time simulations and analysis, thus speeding up decision-making processes. Second, it enhances privacy protection by sharing processed insights rather than raw industrial data, which minimizes the risk of sensitive information being exposed. Additionally, DTs support intelligent industrial networks by providing a comprehensive and contextualized view of operations, which helps in optimizing processes and improving operational efficiency.

Leveraging DT technology for industrial data sharing allows industries to move beyond simple data transfer, enabling more efficient, secure, and insightful data-sharing practices. This approach not only enhances operational performance but also supports better decision-making by providing actionable insights derived from detailed simulations and analyses. Driven by the capabilities of DT in terms of industrial data sharing, the authors in \cite{dietz2019distributed} propose a framework that leverages DT technology to enhance secure data sharing in industrial ecosystems. While DT significantly advances digitization, it introduces new challenges in IT security across industries, particularly during the exchange of DT data between non-trusting parties. For example, synchronizing tasks between DTs, such as those of a power plant, must ensure data integrity to prevent manipulated operations, and maintain confidentiality to protect sensitive information. Their approach combines DT technology with distributed ledger technology to address industrial IT security challenges by ensuring data integrity and confidentiality and eliminating the need for a trusted third party. The framework facilitates secure DT industrial data sharing across an asset's life cycle, offering a robust and efficient solution for industrial applications. In \cite{huang2020blockchain}, the DT significantly enhances data sharing in product life cycle management by leveraging blockchain technology. The DT provides a real-time, high-fidelity virtual representation of the physical product, ensuring that all participators have access to the most current industrial data. By combining two emerging technologies, namely DT and blockchain, a peer-to-peer network is established, which enhances industrial data-sharing efficiency through direct, decentralized exchanges without the need for centralized intermediaries. This approach also secures data storage with cryptographic methods, controlling access to ensure that only authorized participators can view or modify the data. Additionally, the immutable nature of blockchain \cite{nguyen2020blockchain} guarantees the authenticity and integrity of the DT data, as all changes are recorded transparently. Smart contracts further streamline and automate data-sharing processes by executing predefined actions, thereby improving overall efficiency and reliability. This combination of DT and blockchain technologies ensures that data sharing across industries is secure, efficient, and trustworthy.

Various kinds of engineering software and digitalized equipment are widely utilized throughout the lifecycle of industrial products, resulting in the generation of massive amounts of diverse data. However, this data often remains isolated and outdated, leading to inefficiencies and underutilization of valuable information. Traditionally, simulation based on theoretical and static models has been a powerful tool for verification, validation, and optimization during the early planning stages of a system. However, these simulations are rarely applied during system run-time. With the advent of new-generation information and digitalization technologies, the ability to collect more comprehensive data has improved significantly. This has highlighted the need for a method to deeply leverage all available data, leading to the rapid development and increasing interest in the concept of the DT \cite{liu2021review}. Industrial data sharing in DT technology involves the continuous exchange and integration of data between physical assets and their digital counterparts. This process is crucial for creating accurate and dynamic virtual models that reflect the real-time state of physical systems, enabling enhanced monitoring, diagnostics, and decision-making. However, DTs are often misconceived as being equivalent to digital models or digital shadows, yet they represent a more advanced and integrated concept. DTs significantly surpass digital models and digital shadows due to their fully integrated bi-directional data interaction with physical entities \cite{wu2021digital}. Unlike digital models, which lack self-driven data interaction, and digital shadows, which only receive data in a unidirectional manner, DTs facilitate continuous, two-way data exchange. This dynamic interaction enables real-time monitoring, advanced predictive maintenance, and proactive optimization of physical assets. DTs not only reflect the current state of the physical system but also influence and enhance its performance by implementing real-time adjustments\cite{singh2021digital}. This comprehensive, up-to-date representation leads to greater accuracy and reliability, sophisticated scenario analysis, and seamless integration within industrial ecosystems. The continuous feedback loop promotes ongoing improvements and innovation, making DTs an indispensable tool for enhancing efficiency, productivity, and competitive advantage in various industrial applications. DT can also be integrated with other emerging concepts such as 'digital thread', 'digital model' etc. One of the examples of such synchronization can be the study in \cite{pang2021developing}. This study offers an overview of the current state-of-the-art DT and digital thread technology in industrial operations. Both DT and digital thread technologies are transformative, offering significant advantages in enhancing the efficiency of current design and manufacturing processes. DT plays a crucial role in the Industry 4.0 digitalization journey; however, the vast amounts of data generated and collected by DT present challenges in data handling, processing, and storage. The paper presents a new framework that integrates DT with the digital thread to improve data management. The framework is designed to drive innovation, optimize production processes and performance, and ensure information continuity and traceability. It includes components for behavior simulation and physical control, which leverage the connectivity between the twin and thread for seamless information flow and exchange. The framework also outlines specifications for organizational architecture, security, user access, databases, and both hardware and software requirements. It is expected to enhance the optimization of operational processes and information traceability in physical environments, particularly within Industry Shipyard 4.0. Additionally, small and medium-sized enterprises (SMEs) frequently struggle with data management complexities due to diverse databases and insufficient data processing systems. These challenges arise from the diverse formats and sources of data that SMEs must integrate and analyze, leading to inefficiencies and inaccuracies. Many SMEs still rely on manual data acquisition and processing methods, which are time-consuming and prone to errors. Additionally, the lack of robust data processing systems exacerbates these issues, limiting the effectiveness of data-driven decision-making. Consequently, SMEs struggle to adopt fully automated solutions that could streamline data handling and enhance operational efficiency. This fragmentation and inefficiency in data management hinder SMEs' ability to leverage advanced technologies like DTs for real-time insights and optimization. In an attempt to mitigate these issues, \cite{uhlemann2017digital} focuses on advancing automated data acquisition and sharing in SMEs by leveraging DT technology. It highlights the challenges SMEs face with heterogeneous databases and inadequate data processing systems, which hinder the adoption of fully automated data solutions. The paper proposes a learning factory-based approach that upgrades existing systems with multi-modal data acquisition and a flexible, service-oriented optimization environment. By demonstrating the benefits of real-time data sharing and simulation through DT, this approach aims to enhance SMEs' capabilities in managing and utilizing production data effectively, facilitating a more integrated and efficient cyber-physical production system.

\subsection{Data Offloading}
\begin{figure*}[ht!]
     \centering
     \includegraphics[width=0.99\textwidth]{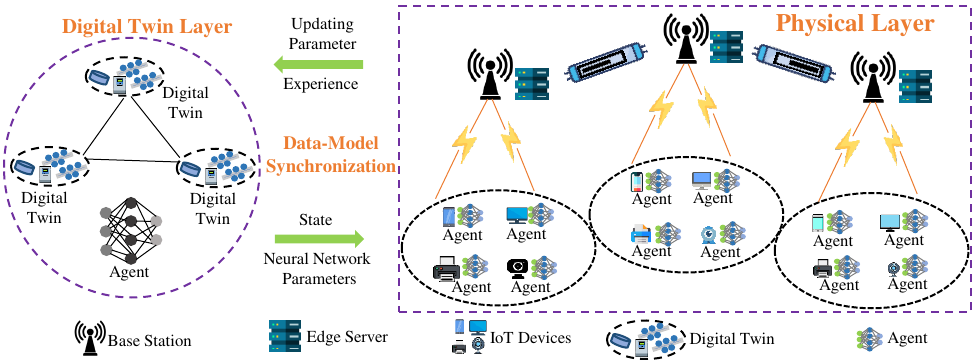} 
     \caption{Dynamic task offloading for digital-twin empowered mobile edge computing networks \cite{chen2023dynamic}.}
     \label{data_offloading_scheme}   
\end{figure*}

DTs in industrial settings have revolutionized data management and operational efficiency, particularly through their impact on industrial data offloading. DTs create accurate virtual replicas of physical assets and processes, enabling real-time monitoring, predictive maintenance, and optimization of industrial operations. This transformation is largely driven by industrial data offloading, which involves transferring vast amounts of industrial data from on-site machinery and sensors to centralized cloud platforms or edge devices for processing and analysis. This process not only alleviates the computational burden on local systems but also enhances data accessibility and analytical capabilities. In practice, data offloading within the DT framework allows industries to manage and process large datasets more efficiently \cite{peng2022distributed}. By shifting the data processing load to more capable cloud or edge computing resources, local systems are freed from intensive computational tasks, leading to improved system performance and responsiveness. This is particularly important in industrial environments where timely and accurate data analysis is critical for maintaining operational continuity and efficiency. Furthermore, the integration of DTs with advanced industrial data offloading techniques ensures that industries can harness the full potential of big data and IIoT. This synergy enables more sophisticated data analytics, supporting improved decision-making processes, reducing downtime through predictive maintenance, and ultimately increasing productivity. A dynamic task offloading scheme for DT empowered mobile edge computing networks in \cite{chen2023dynamic} is presented in Fig. ~\ref{data_offloading_scheme}. For instance, real-time data from physical assets can be offloaded to cloud servers where advanced algorithms analyze the data to predict equipment failures before they occur, allowing for timely interventions and minimizing operational disruptions. As DTs continue to evolve, their integration with efficient industrial data offloading strategies will be crucial for driving the next wave of industrial innovation. The ability to seamlessly manage and process large volumes of data will enable industries to adopt more complex and intelligent systems, fostering an environment of continuous improvement and technological advancement. This evolution will not only enhance the operational capabilities of industries but also contribute to their long-term sustainability and competitiveness in the rapidly changing industrial landscape.

Optimizing industrial data offloading within DT frameworks is necessary to enhance computational efficiency, reduce latency, and improve real-time decision-making capabilities. Hence, the work in \cite{dai2020deep} presents a comprehensive study on the optimization of data offloading within DT frameworks in industrial environments, highlighting the creation and management of virtual replicas of physical assets and processes. The focus is on how data offloading, which transfers substantial volumes of data from on-site machinery and sensors to centralized cloud platforms or edge devices for advanced processing, mitigates local computational loads while enhancing data accessibility and analytical capabilities. By integrating DTs with cutting-edge data offloading techniques, industries can leverage the full potential of big data and IIoT for improved decision-making, predictive maintenance, real-time monitoring, reduced downtime, and increased productivity. The paper introduces novel algorithms and frameworks for optimizing data offloading, supported by detailed performance analysis and practical case studies. This research elucidates the symbiotic relationship between DTs and data offloading, offering innovative solutions that drive the next wave of industrial innovation. Moreover, the authors in \cite{liu2021digital1} introduce an adaptive DT framework for a UAV-enabled MEC network involving multiple mobile terminal users (MTUs), a UAV with a MEC server, multiple resource devices, and a BS. The framework aims to minimize overall system energy consumption by jointly optimizing MTU association, UAV trajectory, transmission power distribution, and computation capacity allocation. By incorporating user mobility and utilizing device-to-device communication links for task offloading, the framework accurately predicts network states and manages computing resource assignments. The paper addresses the complex mixed integer nonlinear optimization problem of energy-efficient MTU association and resource allocation, presenting transformations to reformulate it into a tractable form and achieve near-optimal solutions with reduced complexity. Additionally, a deep reinforcement learning (DRL) approach is proposed to determine MEC offloading decisions, coupled with an iterative algorithm to optimize computation capacity. The study evaluates the effectiveness and superiority of the proposed method through extensive numerical results, demonstrating significant reductions in system energy consumption and providing low-complexity design guidelines for MTUs to efficiently complete tasks. Furthermore, in order to reduce average offloading latency, offloading failure rates, and service migration rates, the study in \cite{sun2020reducing} explores the role of DT edge networks in optimizing data offloading for 6G networks. Recognizing the complexities and unpredictability of MEC environments in 6G, the authors propose leveraging DTs to enhance the accuracy of edge server state estimation and provide training data for offloading decisions. A mobile offloading scheme within DI edge networks aims to minimize offloading latency while managing the costs associated with service migration during user mobility. The study utilizes Lyapunov optimization to transform the long-term migration cost constraint into a multi-objective dynamic optimization problem, which is then addressed using Actor-Critic DRL. This approach significantly reduces average offloading latency, offloading failure rates, and service migration rates compared to benchmark schemes, while also lowering system costs through the assistance of DTs. In industry, automation is crucial for enhancing efficiency, reducing operational costs, and maintaining high levels of precision in manufacturing processes. The integration of MEC with DT technology further optimizes these benefits by providing real-time data processing and simulation capabilities. This combination enables quicker decision-making, minimizes downtime, and improves overall system performance by allowing for intelligent task offloading and better resource management. As industries increasingly rely on interconnected systems and IIoT devices, this approach is essential for staying competitive and meeting modern operational demands. Besides, latency optimization in DT systems is crucial to ensure real-time responsiveness and efficiency in industrial automation and IIoT applications. Hence, the paper in \cite{zhao2023digital} proposes an innovative DT-assisted intelligent partial offloading scheme for Vehicle Edge Computing (VEC), addressing the challenges of dynamic network topology and strict low-delay constraints. By integrating DT technology with improved clustering algorithms and DRL, the paper aims to optimize task offloading decisions and reduce system costs. The scheme features a feedback mechanism to enhance cooperation between digital and physical spaces, creating a closed loop of prediction, offloading, and feedback. The paper demonstrates that the proposed approach effectively reduces computational costs and delays and improves offloading success rates in VEC systems. Another research work with a similar objective is \cite{do2022digital}, which investigates an MEC architecture enhanced by DT for industrial automation, focusing on intelligent industrial data offloading by multiple IIoT devices to multiple MEC servers to reduce end-to-end latency. The study begins by proposing and formulating a practical end-to-end latency minimization problem within a DT-assisted MEC model, considering quality-of-service constraints and computation resources of IIoT devices and MEC servers in industrial IIoT networks. The joint optimization method is proposed, by considering both transmit power of IIoT devices and user association for enhancing system latency savings, highlighting the potential of DT-assisted MEC in optimizing data offloading for industrial applications.

\subsection{Integrated Sensing and Communication}
Integrated sensing and communication (ISAC) within DT services is poised to revolutionize industrial applications by providing unparalleled industrial data integration and operational efficiency. DTs, which create precise virtual replicas of physical assets, machines, and processes, enable real-time monitoring, predictive maintenance, and optimization of industrial operations. The fusion of ISAC into this ecosystem brings a new dimension to how industries manage and utilize their data, fostering a seamless and continuous flow of information between the physical and digital worlds \cite{wei2024integrated}. ISAC enhances DT services by combining advanced sensing capabilities with robust communication technologies. This integration allows for the continuous and accurate transfer of industrial data from physical entities to their digital counterparts. Sensors embedded in machinery and equipment gather extensive real-time data on performance, condition, and environment. This data is then communicated instantly to the DTs, where it is processed and analyzed \cite{he2023physics}. The result is a dynamic, real-time representation of the physical asset that can be used to predict failures, optimize performance, and schedule maintenance, thereby significantly reducing downtime and operational costs. The efficiency of industrial data offloading plays a critical role in this process. By leveraging ISAC, data collected by sensors is transferred to centralized cloud platforms or edge devices for processing and analysis \cite{ding2024joint}. This not only alleviates the computational burden on local systems but also ensures that the data is accessible for immediate analysis and action. The integration of ISAC with DTs means that large volumes of data can be managed more effectively, enabling industries to harness the full potential of big data and IIoT. Advanced analytical tools can process this data to uncover insights, optimize processes, and enhance decision-making. Furthermore, ISAC ensures reliability and speed in data transfer, which is essential for real-time applications. The synchronized sensing and communication framework allows for rapid detection and response to changes in the physical environment. For instance, in manufacturing, ISAC can help maintain optimal production conditions by continuously monitoring equipment performance and environmental factors. Any deviations from the norm can be quickly identified and addressed, minimizing the risk of production delays or quality issues \cite{he2018surveillance}.

Enhancing spectrum efficiency in ISAC within DT services is crucial for optimizing data throughput and reducing interference, which enables more effective real-time monitoring and analysis in industrial applications. Hence, the authors in \cite{cui2023physical} explores advancements in ISAC within DT services, focusing on enhancing physical layer technologies to optimize hardware and spectrum usage. By conducting a comprehensive survey on the current state of DT combined with communication and sensing, the authors identify key challenges and opportunities for improving integration. The study addresses the degree of freedom (DoF) problem in communication and sensing systems, proposing a refined DoF definition specific to sensing systems. To enhance spectrum efficiency, the paper introduces an iterative optimization framework designed to manage the coexistence of communication and sensing functions within DT systems. Additionally, the authors propose a novel waveform design method based on DoF completion to achieve optimal integration gains and minimize mean square error. Moreover, virtualization and resource management in the 6G network are challenging in industries due to the need for efficient handling of diverse, dynamic demands and limited resources, which DT systems address by providing real-time, accurate simulations and optimization of network resources and service provisioning. With a view to addressing the challenges of virtualization and resource management in the 6G network, the work in \cite{gong2022resource} proposing a comprehensive network virtualization architecture that combines DT and network slicing for effective service and user management. In the context of VEC, which faces limitations due to latency-sensitive and computing-intensive applications, the paper introduces an environment-aware offloading mechanism based on ISAC systems. The proposed scheme aims to minimize overall response time by optimizing task scheduling and resource allocation through a Markov decision process. An advanced algorithm integrating Shapley-Q values and deep deterministic policy gradient is employed to solve this optimization problem. Simulation results demonstrate the effectiveness of the proposed approach in enhancing the performance and efficiency of DT services in industrial applications, particularly in managing dynamic vehicle environments and network demands. Furthermore, \cite{li2023adaptive} explores a DT-enhanced integrated sensing, communication, and computation network, where users perform radar sensing and computation offloading on the same spectrum, and UAVs provide edge computing services. It formulates a multi-objective optimization problem to minimize both the beam pattern performance of multi-input multi-output radars and computation offloading energy consumption. By leveraging DT's predictive capabilities, the study addresses estimation deviations and reformulates the problem as a multi-agent Markov decision process, applying a multi-agent proximal policy optimization framework with Beta-policy and attention mechanisms to enhance training performance. This approach effectively balances the trade-off between sensing and computation functions while reducing energy consumption, showcasing the potential of DT in optimizing ISAC systems in industrial applications.



\subsection{Content Caching}
Content caching is a crucial service in modern industries, especially with the rise of IIoT devices and the advent of 6G networks. In industrial environments, where a vast array of IIoT sensors continuously generates industrial data, content caching ensures that frequently accessed information is quickly retrievable without straining network resources \cite{liu2024digital}. For instance, in a smart manufacturing facility, real-time data from production machinery can be cached to provide immediate insights and enable rapid adjustments, reducing latency and improving operational efficiency \cite{xu2022computation}. Similarly, in 6G networks, which are designed to handle massive volumes of data with ultra-low latency, caching mechanisms help manage the high-speed data flow by storing and quickly accessing critical information close to the edge of the network. This reduces the computational burden on central servers and minimizes the need for repeated data retrieval from distant sources. By optimizing data access speed and reducing both latency and network load, content caching enhances real-time decision-making and operational optimization, driving more responsive and efficient industrial processes \cite{yao2023cooperative}. DTs can significantly enhance industrial content caching services by providing a sophisticated framework for managing and utilizing real-time data in industrial settings. By creating accurate virtual replicas of physical assets, machines, processes, and systems, DTs enable dynamic and continuous updates that are crucial for effective content caching. For instance, DTs can monitor and analyze industrial data from IIoT sensors in real-time, identifying patterns and frequently accessed information that can be cached strategically. This predictive capability allows for preemptive caching of relevant data, reducing latency and ensuring that critical information is readily available when needed. In scenarios involving complex industrial processes or large-scale systems, DTs can optimize caching strategies by determining which data is most valuable for immediate access and which can be archived or processed later. Additionally, DTs can manage the synchronization between cached content and its virtual model, ensuring that updates in the physical world are promptly reflected in the cached data. This integration of DTs with content caching not only improves industrial data retrieval speeds but also enhances overall system performance and reliability, leading to more efficient and responsive industrial operations \cite{chen2023resource}. By leveraging the real-time insights and predictive capabilities of DTs, industries can achieve a higher level of operational efficiency and data management.

\begin{figure*}[ht!]
     \centering
     \includegraphics[width=0.99\textwidth]{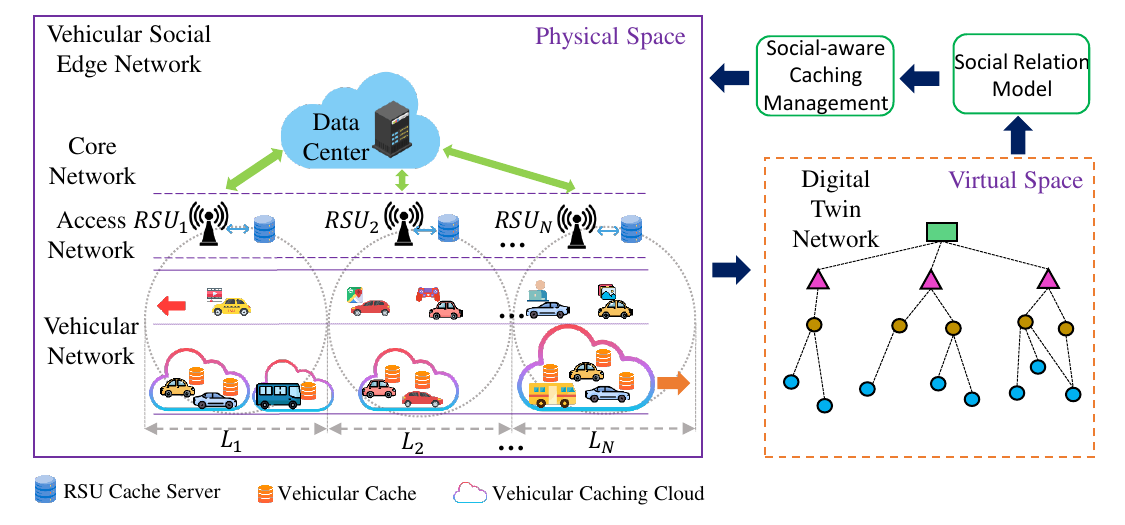} 
     \caption{DT-empowered vehicular social edge network \cite{zhang2021digital}. Enabled by   bidirectional communications between vehicles and virtual entities, DT offers content caching services to support vehicular network management.}
     \label{lat_comp_sub1}   
\end{figure*}

Spurred by the power of DT, the paper in \cite{zhang2021digital} uses DT technology to map the edge caching system into a virtual space, facilitating the construction of a social relation model for optimizing vehicular edge caching. This paper addresses the challenges of efficient content delivery in smart vehicular networks, which face stringent requirements due to the proliferation of powerful applications. It identifies limitations such as constrained storage capacity, limited serving range of individual cache servers, and the highly dynamic topology of vehicular networks. To mitigate these issues, the paper proposes a social-aware vehicular edge caching mechanism that dynamically orchestrates the caching capabilities of roadside units (RSUs) and smart vehicles based on user preference similarity and service availability. The approach leverages DT technology to map the edge caching system into a virtual space, facilitating the construction of a social relation model. This model informs the development of a vehicular cache cloud, which integrates content storage correlations among multiple cache-enabled vehicles in diverse traffic environments. The paper further introduces deep learning-based optimal caching schemes that jointly consider social model construction, cache cloud formation, and cache resource allocation. The proposed schemes are evaluated using real traffic data demonstrating significant advantages in optimizing caching utility. The DT-empowered vehicular social edge network presented by the authors in \cite{zhang2021digital} is illustrated in Fig. ~\ref{lat_comp_sub1}. Moreover, the exponential growth of mobile users and industrial sensor data has strained existing caching resources, making it challenging to meet the escalating data traffic demands \cite{chen2023a3c}. Traditional caching methods across industries often ignore the uplink traffic, which can lead to inefficiencies and poor network performance. Furthermore, emerging delay-sensitive applications such as Virtual Reality, Augmented Reality, and autonomous driving require highly reliable and stable connections to function effectively. These applications cannot tolerate significant delays or interruptions, making optimized caching solutions essential. By dynamically adjusting to changing network conditions and user demands, adaptive industrial caching schemes enhance both the speed of industrial data delivery and the efficiency of energy use, ensuring that IIoT networks can support advanced applications while maintaining robust performance and sustainability. An adaptive caching scheme is introduced in \cite{tan2023adaptive} based on heterogeneous DTs IIoT networks using the evolutionary game theory to optimize both delay and energy consumption in the uplink and downlink. This paper addresses the challenges posed by the massive increase in mobile user and sensor data within heterogeneous IIoT networks, where limited caching resources struggle to meet escalating data traffic demands. It proves the existence of evolutionary stability strategies (ESSs) within the proposed caching scheme and derives expressions relating content popularity to ESS conditions. The proposed approach validates the presence of content evolutionary stability caching strategies, the accuracy of the derived ESS expressions, and the superior caching performance compared to other methods. Optimizing edge caching is pivotal for the advancement of next-generation (nextG) wireless networks, as it ensures high-speed and low-latency services for mobile users. Existing data-driven optimization approaches often fall short by lacking awareness of random industrial data variable distribution, leading to inefficient caching decisions. These methods typically focus on optimizing cache hit rates, neglecting critical reliability concerns such as base station overload and unbalanced industrial cache distribution. Overloaded base stations can suffer severe performance degradation, resulting in increased latency and potential service interruptions. Additionally, unbalanced cache usage causes some caches to be overutilized while others remain underutilized, further straining network resources. These oversights jeopardize network stability and degrade user experience, particularly in real-time applications like autonomous driving and virtual reality. Thus, a more holistic approach to industrial edge caching optimization is necessary, considering both data distribution and network reliability to ensure robust and efficient performance in next-generation (nextG) wireless networks. Hence, the authors in \cite{zhang2024digital} introduce a novel DT-assisted optimization framework, called D-REC, aimed at enhancing edge caching in nextG wireless networks. Recognizing the limitations of existing data-driven approaches, which often neglect the distribution of random data variables and critical reliability concerns, D-REC integrates RL with diverse intervention modules to ensure reliable caching. The framework employs a joint vertical and horizontal twinning approach to create efficient network DTs, which serve as RL optimizers and safeguards. These DTs provide extensive datasets for training and predictive evaluation of cache replacement policies. By incorporating reliability modules into a constrained Markov decision process, D-REC adaptively adjusts actions, rewards, and states to adhere to beneficial constraints, thereby minimizing the risk of network failures. Theoretical analysis shows that D-REC achieves comparable convergence rates to traditional data-driven methods without compromising caching performance. Their proposed method significantly outperforms conventional approaches in cache hit rate and load balancing, while effectively enforcing predetermined reliability intervention modules. To address the challenges of explosive traffic growth and efficient content delivery in the Internet of Everything (IoE) environment, propelled by the rapid development of 5G/6G and AI technologies, the paper in \cite{yi2022digital} proposes a highly efficient content delivery framework by reconstructing the IoE environment as an end-edge-cloud collaborative system enhanced by DT technology. The framework begins with a content popularity prediction scheme using the Temporal Pattern attention-enabled Long Short-Term Memory (LSTM) model to identify critical content. The predicted results inform a caching scheme, leveraging RL to decide optimal locations for sinking this critical content. Additionally, a collaborative routing scheme is introduced to ensure efficient content access, aiming to minimize overhead. 

\subsection{Resource Allocation}
Resource allocation is a critical service in industries, directly impacting operational efficiency, cost management, and service quality. In industrial settings, effective resource allocation ensures that assets such as personnel, equipment, machines, and materials are optimally distributed to meet production demands, maintain service levels, and maximize profitability. Proper allocation involves not just assigning industrial resources to specific tasks but also balancing and optimizing their use to prevent bottlenecks, reduce downtime, and avoid waste. However, resource allocation presents several challenges. Industries often face issues such as resource scarcity, where demand exceeds availability, leading to delays and increased operational costs \cite{guo2023intelligent}. Furthermore, dynamic and unpredictable environments, such as manufacturing floors or service operations, can result in fluctuating demands that complicate the allocation process. Inefficient industrial resource allocation can also lead to underutilization or overutilization of resources, causing imbalances that affect productivity and increase maintenance needs. Additionally, a lack of real-time data and visibility can hinder the ability to make informed allocation decisions, exacerbating these issues. DT technology offers a powerful solution to these challenges by providing a virtual replica of physical systems and processes\cite{cheng2024toward}. By simulating real-world operations, DTs enable industries to monitor resource usage and performance in real time, allowing for more accurate and dynamic resource allocation. DTs can integrate industrial data from various sources, such as sensors, machines, and historical records, to create comprehensive models that predict resource needs and identify potential issues before they arise \cite{zhang2018dynamic}. For instance, in a manufacturing environment, a DT can model production lines and resource flows, providing insights into where resources are most needed and predicting when and where shortages might occur. This allows for proactive adjustments and optimization of industrial resource allocation. In service industries, DTs can track customer interactions and service demand patterns, facilitating better industrial scheduling and staffing decisions. Furthermore, DTs enable scenario planning and simulation, allowing industries to test different allocation strategies and assess their impact before implementation. This reduces the risk of making costly mistakes and ensures that industrial resources are allocated in the most efficient manner possible \cite{tang2023digital}. Overall, by leveraging DT technology, industries can enhance their resource allocation processes, improve operational efficiency, and respond more effectively to dynamic and complex challenges.

DTs can be a highly efficient approach to solve the aforementioned issues and improve industrial resource allocation even in the unpredictable network dynamics that complicate industrial resource supply and demand matching. The work in 
\cite{sun2021dynamic} addresses the problem of efficient resource allocation in the Internet of Vehicles (IoV) empowered by aerial communications where network dynamics can be highly unpredictable. It introduces a dynamic DT of aerial-assisted IoV to effectively manage time-varying resource supply and demand, enabling unified resource scheduling and allocation. The paper designs a two-stage incentive mechanism based on Stackelberg game theory, with the DT of vehicles or roadside units (RSUs) acting as the leader and RSUs providing computing services as the followers. In the first stage, the mechanism determines the computing resources RSUs are willing to offer based on vehicle preferences. To enhance vehicle satisfaction and overall energy efficiency, a distributed incentive mechanism using the alternating direction method of multipliers is proposed. This mechanism optimizes resource allocation for each vehicle and can be executed concurrently at multiple RSUs to reduce delays and computational burdens on UAVs. In \cite{hazarika2023radit}, the study explores the enhancement of resource allocation in IoV networks using DT technology. It considers a scenario where MEC servers are deployed at roadside units (RSUs), with UAVs acting as relays to provide ubiquitous connectivity, even in areas without RSU coverage. By establishing a virtual representation of the IoV network in the aerial network as a DT, the paper captures real-time dynamics to perform efficient resource allocation for delay-intolerant tasks. The proposed intelligent delay-sensitive task offloading scheme leverages local execution, vehicle-to-vehicle (V2V), and vehicle-to-roadside-unit (V2I) offloading modes based on energy consumption. Furthermore, a multi-network DRL-based resource allocation algorithm (RADiT) is introduced to maximize the utility of the IoV network while optimizing task offloading. The performance of RADiT is compared with and without V2V computation mode, and evaluated against the soft actor-critic DRL algorithm and a non-DRL greedy approach. 

Production logistics (PL) is gaining significant attention in industrial supply chain research. The spatial disorder and temporal asynchrony of PL resources, resulting from uncertainty and dynamic conditions, present substantial challenges to efficient resource allocation in industrial settings. The inability to accurately obtain and utilize the spatial-temporal values of PL resources leads to unnecessarily long travel distances and excessive waiting times, hindering the sustainable performance of PL operations. This inefficiency impacts overall productivity and operational costs, underscoring the need for advanced solutions to optimize resource management in the industrial sector. In response to these issues, 
the research in \cite{zhao2022digital} proposes a novel PL industrial resource allocation approach based on the dynamic spatial-temporal knowledge graph (DSTKG) to enhance efficient resource allocation. By analyzing IIoT signal data generated from large-scale deployed IIoT devices through deep neural networks, the study derives spatial-temporal values critical for resource management. The DSTKG model creates a DT replica with spatial-temporal consistency, enabling the reasoning and completion of relationships based on PL task information. Resources are then allocated efficiently using graph algorithms applied to the directed and weighted graph. A case study is conducted to verify the feasibility and practicality of this approach, and the results demonstrate the effectiveness of the proposed methodology in improving PL operations by reducing unnecessary travel distances and waiting times. Apart from unpredictable network dynamics, high vehicle mobility in VEC leads to frequent changes in network topology and connectivity, making it difficult to allocate resources efficiently. These fluctuations cause challenges in maintaining stable connections and optimizing resource distribution among edge nodes. To enhance VEC network performance, a framework has been proposed in \cite{jeremiah2024digital}, which integrates DT technology to create virtual replicas of network nodes for real-time estimation, prediction, and evaluation. This central DT enables collaboration between edge nodes, such as RSUs or small cells eNodeB, and provides real-time resource information. The framework employs channel state information for RSU selection and uses a non-orthogonal multiple access protocol for vehicle communication with RSUs. The aim is to maximize the VEC system computation rate and minimize task completion delay by jointly optimizing offloading decisions, subchannel allocation, and RSU association. The optimization problem is modeled as a Markov decision process and solved using the Advantage actor-critic algorithm.




\subsection{Wireless Networking}

Wireless networking inside an industry is a critical service in modern industries, providing the backbone for seamless communication, industrial data transfer, and real-time monitoring. Its importance is multifaceted, affecting various operational aspects and driving efficiency, productivity, and innovation. In manufacturing, wireless networks enable the implementation of smart factories where machinery, sensors, and control systems communicate without the constraints of wired connections. This facilitates the deployment of IIoT devices, leading to improved process automation, predictive maintenance, and resource management \cite{9429703}. Real-time data collection and analysis are paramount in optimizing production lines, reducing downtime, and enhancing overall equipment effectiveness. In logistics and supply chain management, wireless networking inside an industry ensures that goods can be tracked and managed efficiently throughout their journey. radio frequency identification and GPS technologies, powered by robust wireless networks, provide accurate location and status updates of shipments, reducing losses and enhancing inventory management\cite{10628026}. This real-time visibility is crucial for just-in-time delivery models, where timely and precise information flow can significantly impact industrial operational efficiency and customer satisfaction. Moreover, industries such as healthcare, retail, and transportation rely heavily on wireless networks to support their critical operations. In healthcare industry, wireless networks enable the use of mobile medical devices, telemedicine, and patient monitoring systems, improving patient care and operational efficiency. In retail, wireless technology supports point-of-sale systems, inventory management, and customer engagement strategies, creating a seamless shopping experience. Transportation sectors benefit from wireless networks through enhanced fleet management, traffic monitoring, and autonomous vehicle operations. The flexibility and scalability of wireless networks allow industries to adapt quickly to changing demands and technological advancements. As businesses increasingly adopt digital transformation strategies, the role of industrial wireless networking becomes even more significant. Advanced wireless technologies like 5G and Wi-Fi 6 offer higher speeds, lower latency, and greater capacity, paving the way for more sophisticated applications such as augmented reality (AR), virtual reality (VR), and artificial intelligence (AI) integration \cite{9374645}.

DT technology addresses critical challenges in industrial wireless networking by providing real-time, virtual replicas of machines, and physical network elements \cite{9420037}. These digital models simulate network behavior across industries, enabling proactive management and optimization of industrial network resources. DTs facilitate dynamic network planning and fault detection by predicting potential issues before they impact operations, such as congestion or signal interference. They enhance network performance by analyzing real-time industrial data from network nodes and adjusting configurations accordingly. Through continuous monitoring and simulation, DTs support efficient industrial resource allocation and adaptive network management, ensuring optimal industrial performance even in high-demand scenarios. This proactive approach reduces downtime, improves reliability, and enhances overall network efficiency, making DTs a powerful tool for modernizing and optimizing industrial wireless networking. The study in \cite{zhao2022elite} introduces Intelligent DT-based software-defined vehicular networks to address the limitations of traditional software-defined vehicular networks. The proposed approach enhances routing in vehicular networks through a four-phase process. First, it uses parallel agents within a virtual DT network to learn and generate routing policies. Next, these policies are refined and combined based on complex communication needs. In the deployment phase, the most appropriate policy is selected according to real-time network conditions, and a road path is calculated and communicated to the requesting vehicle. Finally, relay vehicles are chosen for each hop along the selected path. By leveraging DT technology, the proposed approach significantly improves packet delivery ratio, reduces end-to-end delay, and lowers communication overhead compared to traditional methods, showcasing its effectiveness in optimizing vehicular network routing. Moreover, The integration of the IIoT in industrial settings is driving the widespread use of DT technology, particularly at the network edge. This application helps manage the complexity arising from diverse, siloed application solutions and varying protocols from different manufacturing tools and enterprise systems. However, network heterogeneity presents a significant challenge in this context, as it can severely impact the design and deployment of DT-oriented applications. The diverse nature of IIoT systems introduces inconsistencies and interoperability issues that complicate the seamless integration and operation of DTs. Addressing these challenges is crucial for optimizing industrial DT functionality and achieving efficient industrial operations. To tackle these issues, the article in \cite{bellavista2021application} introduces the application-driven DT networking middleware, designed to address two main objectives. First, it simplifies interactions among diverse devices by enabling DTs to use IP-based protocols rather than specialized industrial ones, thus improving packet content expressiveness through standardized data enrichment. Second, it dynamically manages network resources in edge industrial environments by applying software-defined networking techniques. This approach allows for the adaptation of communication mechanisms to better-fit application requirements, ranging from basic IP protocols to more complex systems based on packet content. 
Emerging technologies and applications have made networks increasingly complex and heterogeneous, escalating operational costs and risks. DT networks can alleviate these challenges by providing virtual environments that allow users to simulate and understand the impact of modifications on network performance. This virtual modeling helps in making informed decisions, thereby enhancing efficiency and reducing the risks associated with network operations. Therefore, \cite{hui2022digital} addresses the challenges of evaluating network performance within industrial settings, where the complexity and heterogeneity have escalated due to emerging technologies and applications. It critiques traditional simulation and analytical methods, highlighting their inefficiency, inaccuracy, and inflexibility for ``What-if" performance evaluations, and promotes data-driven methods as more promising alternatives. The paper outlines three key requirements for effective performance evaluation in industrial DT services: fidelity, efficiency, and flexibility. It provides a comparative analysis of various data-driven methods, examining current trends in data, models, and their applications. The paper also identifies conflicts between the models' ability to process diverse inputs and the limited data available from production networks. Furthermore, it explores opportunities for enhancing data collection, model construction, and the practical application of performance models in industrial contexts, aiming to offer a valuable reference for performance evaluation and to drive future research in DT networks for industrial applications. Moreover, network modeling is crucial for optimizing Quality of Service (QoS) in industrial settings \cite{10375691}. Accurate estimation of SLA metrics such as delay and jitter is essential for maintaining reliable and efficient operations. However, current modeling techniques struggle to provide precise estimates in environments with intricate QoS-aware queueing policies, such as strict priority, Weighted Fair Queuing, and Deficit round-robin. These limitations hinder the ability to effectively configure routing and queue scheduling policies, ultimately affecting the overall network performance and reliability. The inability to model these complex interactions accurately poses significant challenges for industries relying on high-performance network operations. The research in \cite{ferriol2022building} introduces a graph neural network-based model, TwinNet, designed to address these challenges. TwinNet, a DT, accurately estimates SLA metrics by understanding the interplay between queueing policies, network topology, routing configuration, and input traffic matrix. It generalizes well to new scenarios, showing high accuracy in estimating end-to-end path delays across various real-world topologies and configurations. This capability enhances SLA-driven network optimization and supports effective "what-if" analyses, making it highly valuable for industrial applications.
\begin{table*}
    \centering
    \caption{Taxonomy of DT services in industries.}
    \label{tab:taxonomy_services}
    \small
    \begin{adjustbox}{max width=\textwidth}
    \begin{tabular}{|m{1cm}|c|m{2.5cm}|m{6cm}|m{5cm}|}
        \hline
        \textbf{Issue} & \textbf{Ref.} & \textbf{Use case} & \textbf{Key contributions} & \textbf{Limitations} \\ \hline
        \multirow{7}{*}{\rotatebox[origin=c]{90}{\makecell{DT for \\ Data Sharing \quad}}} & \cite{dietz2019distributed} & Industrial IIoT data sharing & A secure data sharing framework using distributed ledger technology with DT. & Limited demonstration through a single-use case. \\ \cline{2-5}
        & \cite{huang2020blockchain} & Industrial IIoT data sharing & A data sharing method for DT of product lifecycle based on DT and blockchain. &  Evaluation limited to a specific product lifecycle scenario. \\ \cline{2-5}
        & \cite{pang2021developing} & Industry Shipyard 4.0 data sharing & A better data sharing and management framework for industrial operations using DT and digital thread. & Framework not yet validated in diverse industrial settings. \\ \cline{2-5}
        & \cite{uhlemann2017digital1} & Cyber-physical production systems data sharing & A multi-modal data sharing and acquisition approach for production systems implemented by DT. & Limited focus on small and medium-sized enterprises. \\ \hline
        
        \multirow{8}{*}{\rotatebox[origin=c]{90}{\makecell{DT for \\ Data Offloading \quad}}} & \cite{dai2020deep} & IIoT data offloading & An energy-efficient DT scheme for IIoT stochastic data task offloading. & Focuses on theoretical optimization without extensive real-world validation. \\ \cline{2-5}
        & \cite{liu2021digital} & Edge data offloading & A DT-assisted task offloading scheme using edge collaboration, CSI, and blockchain. & Computational complexity due to dual optimization models and algorithms. \\ \cline{2-5}
        & \cite{sun2020reducing} & Mobile data offloading & A mobile offloading scheme to minimize latency and migration costs using DTs and deep reinforcement learning. & Results based on simulations may not capture all real-world complexities. \\ \cline{2-5}
        & \cite{zhao2023digital} & Vehicle edge data offloading & An intelligent partial data offloading scheme using DT and DRL. & Feedback mechanism effectiveness might vary with different network conditions. \\ \cline{2-5}
        & \cite{do2022digital} & IIoT data offloading & An intelligent task offloading scheme for industrial IIoT networks based on DT and iterative optimization. & Performance based on simulations with conventional methods as benchmarks. \\ \hline
        
        \multirow{3}{*}{\rotatebox[origin=c]{90}{\makecell{DT for Integrated \quad \quad \\ Sensing and \quad \quad \\ Communication \quad \quad}}} & \cite{cui2023physical} &  Edge network & A DT-enabled iterative optimization framework and new waveform design with ISAC. & Based on simulations, which may not capture all real-world complexities. \\ \cline{2-5}
        & \cite{gong2022resource} & Edge network & A holistic network visualization framework using DT with ISAC &  Potential challenges and difficulties in implementing into existing real-world infrastructure. \\ \cline{2-5}
        & \cite{li2023adaptive} & UAV-assisted edge computing environments & A DT-enhanced ISAC scheme with multi-agent optimization to minimize energy consumption. & Data privacy should be considered in such a scheme. \\ \cline{2-5}
        & \cite{ding2024joint} & Vehicular network & A DT-enhanced ISAC approach for optimizing vehicle assignment and beamforming. & The proposed framework is not scalable. \\ \hline
                
        \multirow{3}{*}{\rotatebox[origin=c]{90}{\makecell{DT for Content Caching \quad \quad}}} & \cite{zhang2021digital} &  Vehicular edge caching & A a DT-enhanced social-aware vehicular edge caching mechanism with deep learning for optimizing caching utility and resource allocation. & Evaluation based on real traffic data; may not account for all practical deployment scenarios. \\ \cline{2-5}
        & \cite{tan2023adaptive} & IIoT content caching & An adaptive DT-based caching scheme using evolutionary game theory for optimizing delay and energy consumption in heterogeneous IIoT networks. &  Potential challenges and difficulties in implementing into existing real-world infrastructure. \\ \cline{2-5}
        & \cite{zhang2024digital} & nextG wireless networks edge caching & A DT-assisted optimization framework integrating RL for reliable edge caching in nextG networks with improved cache hit rate and load balancing. &  Potential challenges and difficulties in maintaining user privacy. \\ \cline{2-5}
        & \cite{yi2022digital} & 5G/6G IoE environments edge content caching & A DT-enhanced content delivery framework using LSTM for content prediction and RL for optimal caching and routing in IoE environments. & Reliance on predictive models may not account for real-time traffic dynamics and unexpected content demands. \\ \hline
        
    \end{tabular}
    \end{adjustbox}
\end{table*}




\begin{table*}[ht!]
    \centering
    \caption{Taxonomy of DT services in industries (continued).}
    \label{tab:taxonomy_services_continued}
    \small
    \begin{adjustbox}{max width=\textwidth}
    \begin{tabular}{|m{1cm}|c|m{2.5cm}|m{6cm}|m{5cm}|}
        \hline
        \textbf{Issue} & \textbf{Ref.} & \textbf{Use case} & \textbf{Key contributions} & \textbf{Limitations} \\ \hline
        \multirow{3}{*}{\rotatebox[origin=c]{90}{\makecell{DT for Metaverse \quad \quad \quad \quad}}} & \cite{aloqaily2022integrating} &  Industrial metaverse & A framework integrating DTs with 6G, blockchain, and AI to enhance performance, security, and scalability of metaverse services. & Implementation challenges and integration complexities with emerging technologies. \\ \cline{2-5}
        & \cite{han2022dynamic} & Industrial metaverse & A DT-enabled framework to enhance synchronization and quality control for reliable data exchange and operational harmony within the metaverse.  &  Concentrates on DT synchronization and quality control, potentially overlooking broader metaverse integration issues. \\ \cline{2-5}
        & \cite{jagatheesaperumal2023semantic} & Industrial network & A scheme to ensure contextually accurate and relevant information by linking all devices to a semantic model, enhancing coordination and decision-making in the metaverse. & Focuses on semantic accuracy and coordination, which may not address broader scalability issues in complex metaverse applications. \\ \cline{2-5}
        & \cite{kulkarni2024hybrid} & Metaverse in healthcare industry & A hybrid model leveraging DT technology for robust disease prediction and incremental learning in the metaverse. & May not fully address the integration of disease prediction with other metaverse healthcare applications or broader DT functionalities. \\ \hline

        \multirow{3}{*}{\rotatebox[origin=c]{90}{\makecell{DT for Resource Allocation \quad \quad}}} & \cite{sun2021dynamic} &  Aerial-assisted IoV networks & A dynamic DT-based resource allocation framework in IoV using a two-stage Stackelberg game theory mechanism to optimize computing resource scheduling. & Effectiveness of the mechanism may vary with network unpredictability. \\ \cline{2-5}
        & \cite{hazarika2023radit} & IoV networks & A DT-based intelligent task offloading scheme using RADiT for efficient resource allocation in IoV networks, incorporating V2V and V2I modes. &  Performance comparison limited to specific scenarios; may not generalize to all IoV environments. \\ \cline{2-5}
        & \cite{zhao2022digital} & IIoT networks & A DT-based PL resource allocation approach that enhances efficiency through spatial-temporal consistency and graph algorithms. &  Effectiveness of the approach may be limited by the accuracy of IIoT signal data and model assumptions. \\ \cline{2-5}
        & \cite{jeremiah2024digital} & VEC network & A DT-based framework for real-time optimization of VEC network performance, integrating offloading decisions, resource allocation, and RSU association. & Solution performance may depend heavily on the accuracy of channel state information and the complexity of the Markov decision process model. \\ \hline

        \multirow{3}{*}{\rotatebox[origin=c]{90}{\makecell{DT for Wireless networking \quad \quad}}} & \cite{zhao2022elite} &  Vehicular wireless network routing & An Intelligent DT-based approach for optimizing vehicular network routing, enhancing packet delivery, reducing delay, and lowering communication overhead. & Performance improvements may be constrained by the accuracy of DT simulations and real-time network conditions. \\ \cline{2-5}
        & \cite{bellavista2021application} & Industrial IIoT and edge networks & A DT networking middleware that simplifies device interactions and dynamically manages network resources using IP-based protocols and software-defined networking techniques. &  Middleware effectiveness may vary based on the complexity of network environments and specific application requirements. \\ \cline{2-5}
        & \cite{hui2022digital} & Industrial networks & A DT-based data-driven approach for network performance evaluation in industrial services.  &  Challenges in reconciling diverse input processing capabilities of models with the limited data available from production networks. \\ \cline{2-5}
        & \cite{ferriol2022building} & Industrial SLA-driven network o & A graph neural network-based DT model for accurate SLA metric estimation and end-to-end path delay prediction in various network topologies. & Generalization accuracy may vary with highly complex or previously unseen network configurations. \\ \hline
        
    \end{tabular}
    \end{adjustbox}
\end{table*}

\subsection{Metaverse}
The metaverse represents a transformative advancement in digital interaction and experience, offering significant potential benefits across various industries. As a convergence of VR, AR, and other immersive technologies, the metaverse provides a comprehensive platform for businesses to innovate, engage, and optimize their operations \cite{9865226}. In industries such as manufacturing, the metaverse enables the creation of DTs—virtual replicas of physical assets, machines, actuators, processes, or environments. These DTs facilitate real-time monitoring, simulation, and analysis, allowing for enhanced predictive maintenance, operational efficiency, and streamlined design processes \cite{10017413}. By integrating VR and AR technologies, engineers and designers can interact with 3D models and prototypes in a virtual space, making adjustments and testing scenarios without the constraints of physical limitations. Retail and e-commerce benefit from the metaverse through immersive shopping experiences. Virtual stores and showrooms allow customers to explore products in a 3D environment, try virtual fittings, and interact with digital sales representatives. This not only enhances customer engagement but also reduces the need for physical inventory and store space, leading to cost savings and a more sustainable business model. In training and education, the metaverse offers immersive learning environments where employees can undergo realistic simulations of complex procedures and scenarios. This approach improves training outcomes by providing hands-on experience in a risk-free setting, enhancing skills retention, and reducing training costs. For example, medical professionals can practice surgeries using virtual simulations, while corporate teams can engage in virtual team-building exercises and workshops. Furthermore, the metaverse can revolutionize collaboration and remote work \cite{9786719}. Virtual offices and meeting spaces enable teams to interact as if they were physically present, fostering better communication and collaboration across geographical boundaries. This not only enhances productivity but also supports flexible working arrangements and reduces travel-related expenses. However, metaverse in industrial settings presents several challenges, including managing the massive amounts of data generated by immersive simulations, ensuring real-time synchronization between virtual and physical environments, and maintaining system security and integrity \cite{10517506}. The complexity of integrating diverse IIoT devices and sensors into a cohesive virtual model adds to these difficulties, potentially leading to data inconsistencies and inefficiencies. 

DT technology can manage these challenges by creating dynamic, virtual replicas of physical assets and processes. DTs enable real-time monitoring and simulation of industrial operations, providing accurate and synchronized virtual models that reflect the current state of physical systems. This helps in identifying discrepancies between the virtual and real worlds, optimizing resource allocation, and predicting potential issues before they impact operations. Additionally, DTs enhance security by allowing for continuous testing and validation of virtual environments against potential threats. Propelled by the advantages of DT, research work documented in \cite{aloqaily2022integrating} explores the challenges and requirements for realizing advanced metaverse services, focusing on integrating DT technology with cutting-edge innovations like 6G networks, blockchain, and AI. It highlights the limitations of traditional processing, communication, and storage technologies in supporting the scalability and user experience needed for immersive metaverse environments. The paper proposes a comprehensive framework that combines DTs with these technologies to ensure continuous and reliable end-to-end metaverse services. By leveraging DTs to create real-time virtual replicas of physical environments and integrating them with 6G networks, blockchain, and AI, the framework aims to enhance the performance, security, and scalability of metaverse applications. The article also outlines the key requirements for implementing this integrated approach and offers insights into the future developments of metaverse technologies. Moreover, accurately modeling both tangible and intangible aspects of the Metaverse is essential for creating a comprehensive and realistic virtual environment. Understanding complex social relations through DTs ensures that interactions and behaviors in the Metaverse mirror real-world dynamics, enhancing the authenticity and effectiveness of virtual experiences. Driven by the potential of DT, the authors in \cite{lv2022building} review how DTs can be utilized to model both tangible and intangible aspects of the Metaverse, including various scales and states of objects, as well as complex social relations. They introduce principles and laws like broken windows theory, small-world phenomenon, survivor bias, and herd behavior to guide the construction of DT models for social interactions within the Metaverse. The review explores how these models can effectively map real-world entities and relationships to their virtual counterparts in the Metaverse. In industrial settings, accurate and real-time data from DTs are vital for optimizing processes, improving decision-making, and enhancing productivity. However, the challenge of ensuring high-quality shared DTs across various virtual service providers (VSPs) can lead to inconsistencies and inefficiencies. By proposing a dynamic hierarchical framework and advanced game-theoretic models, the paper in \cite{han2022dynamic} offers a solution to enhance the synchronization and quality control of DTs. This, in turn, ensures reliable data exchange and operational harmony, thereby driving innovation, reducing costs, and improving overall performance in industrial applications. The problem addressed in this paper is crucial for industries because it tackles the interoperability and quality assurance of DTs in the metaverse, which are essential for seamless integration and efficient operation of virtual services. The ability to leverage shared DTs effectively across multiple VSPs can lead to more collaborative and scalable industrial ecosystems, fostering advancements in automation, predictive maintenance, and smart manufacturing. 

Moreover, it is necessary to build a semantic model in order to achieve accurate and task-oriented information extraction by linking all devices, machines in the Metaverse environment for faithful message interpretation. Without a unified semantic model, devices, machines, and systems may interpret messages inconsistently, leading to miscommunications and operational inefficiencies. By linking all devices and machines to a semantic model, the framework in \cite{jagatheesaperumal2023semantic} ensures that information is contextually accurate and relevant to specific tasks, thereby enhancing coordination and decision-making. This precise interpretation of messages is essential for maintaining seamless interactions and optimizing performance in dynamic and interconnected metaverse applications, particularly in complex industrial settings where real-time accuracy and reliability are paramount. Furthermore, integrating DT with the metaverse in the healthcare industry offers a transformative approach to patient care and disease management. DTs create detailed virtual replicas of patients, enabling real-time monitoring and personalized treatment plans within a simulated environment. This integration facilitates advanced predictive modeling and scenario testing, enhancing early disease detection and intervention strategies. By merging DTs with metaverse platforms, healthcare providers can simulate complex medical scenarios and improve decision-making processes. Additionally, it supports immersive training and education for medical professionals, ultimately leading to better patient outcomes and more efficient healthcare delivery. The work detailed in \cite{kulkarni2024hybrid} introduces DAE-BLS, a hybrid model combining Denoising AutoEncoder (DAE) with Broad Learning System (BLS), to enhance disease prediction within the context of consumer health in the metaverse. By leveraging DT technology and comprehensive medical data, DAE-BLS addresses challenges like gradient instability and slow training faced by traditional models, offering robust feature extraction and high accuracy for disease prediction. The model's capability to adapt quickly through incremental learning is particularly valuable for dynamic healthcare scenarios in the metaverse.
\begin{figure*}[h]
     \centering
     \includegraphics[width=\linewidth]{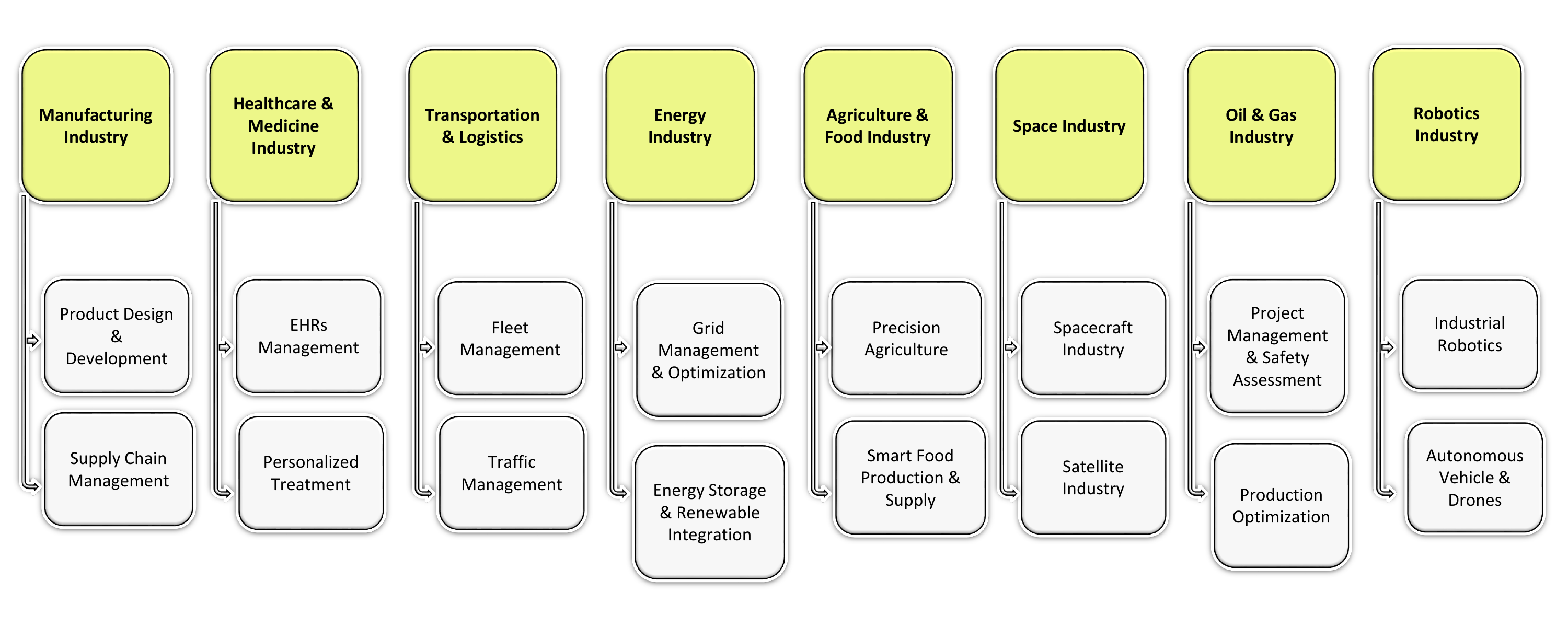}
     \caption{  Summary of DT application domains presented in this paper.}
     \label{Application_domains}   
\end{figure*}
\section{DT Applications in Industries} \label{SectionIV}

In this section, we present a comprehensive overview on the applications of DT in a wide range of domains, including manufacturing industry, healthcare and medicine industry, transportation and logistics, energy industry, agriculture and food industry, space industry, oil and gas industry, and robotics industry as well as the applied use case domains, which are summed up in Fig. \ref{Application_domains}. Additionally, we summarize the taxonomy of DT applications across industries in Table ~\ref{tab:DT_applications} continued to Table ~\ref{tab:DT_applications_continued}. 

\subsection{Manufacturing Industry}
DT technology is profoundly reshaping the manufacturing industry by driving substantial improvements in operational efficiency, predictive maintenance, and the optimization of production processes. This dynamic interaction between the physical and digital realms allows manufacturers to fine-tune production processes, reduce waste, and respond rapidly to changes in demand or production conditions. For example, in assembly lines where product variations and customization are common, DTs allow for real-time adjustments that ensure consistency in quality and efficiency \cite{liu2021digital}. DTs also play a pivotal role in enhancing predictive maintenance strategies. In industries such as automotive manufacturing, where machinery downtime can result in significant financial losses, the ability to predict and prevent failures is invaluable \cite{aivaliotis2019use}.
Furthermore, DTs are revolutionizing supply chain management within the manufacturing sector. It allows for the optimization of supply chain operations, enabling manufacturers to anticipate disruptions, evaluate the impact of various scenarios, and make informed decisions to mitigate risks \cite{grieves2022intelligent}. For instance, during times of global supply chain disruptions, as was the case with the COVID-19 pandemic, DTs provided manufacturers with the agility to reconfigure supply networks and maintain operational continuity \cite{lv2022digital}. Reviewing the literature in the DT-based manufacturing domain, we here focus on analyzing the applications of DT in product design and development as well as supply chain management.
\subsubsection{DT for Product Design and Development}
In product design, DTs play a crucial role in addressing the challenges associated with isolated and fragmented lifecycle data. Traditional approaches often focus on physical products, leaving virtual models and data poorly integrated. DTs bridge this gap by converging physical and virtual data, enhancing the efficiency and intelligence of product design processes. Recent studies highlight the potential of DTs to transform product lifecycle management by integrating cyber-physical data streams \cite{tao2019digital}. In this case, physical and digital entities of the DT framework exchange data seamlessly by leveraging wireless communication technologies. This two-way communication supports more informed decision-making and drives smarter, more sustainable design practices. The introduction of new frameworks and methods demonstrates how DTs can streamline product development, offering real-time updates and continuous optimization across design, manufacturing, and service stages. However, the focus remains predominantly on physical data analysis, with virtual models often overlooked. To address this, the research in \cite{tao2019digital} presents a novel approach to product design utilizing DT technology, starting with a review of product design evolution. It then outlines and examines the framework for DT-driven product design (DTPD), incorporating wireless communication to facilitate real-time interaction between physical products and their digital counterparts. A case study is included to demonstrate the practical application and benefits of this new method in product design. 

Meanwhile, the author in \cite{lin2021evolutionary} focused on the growing need for enhanced flexibility and adaptability to develop industrial products that can adapt to increasingly complex and dynamic environments. Traditional DTs provide deterministic feedback for behavior adjustments, which can limit their adaptability. To overcome these limitations, this paper introduces an advanced concept called the Evolutionary DT (EDT). The EDT framework integrates supervised learning to create a more accurate virtual model of the physical world. This approach aims to enhance the adaptability and flexibility of industrial products by wirelessly exchanging data between physical assets and their digital twins, allowing for faster decision-making. This model then supports the exploration of optimal solutions through reinforcement learning, conducted in multiple virtual environments.  \lq DT\rq\ has emerged as a promising solution shown in another study \cite{howard2019digital}, where the authors explore the potential of DTs for virtual validation, discusses their role in accelerating product development, and highlights the key challenges that need to be addressed for successful implementation.

\subsubsection{DT for Supply Chain Management}
In the rapidly evolving supply chain management landscape, the need for real-time data analysis and predictive insights has become increasingly critical. Traditional methods often fall short in addressing the complexities and dynamic nature of modern supply chains. Some research tried to address these challenges by leveraging DT technology, offering a sophisticated approach to optimize supply chain processes through virtual simulations and data-driven decision-making. For example, in \cite{marmolejo2020design}, the authors focused on designing and developing a DT specifically for a pharmaceutical company's supply chain. The virtual replica of the company's processes sends data wirelessly through the cellular network in order to simulate and analyze the behavior of these processes to enhance their effectiveness. The DT model was intended to provide insights into past performance, optimize current operations, and predict future outcomes, thereby making business processes more robust. Meanwhile, in \cite{park2021architectural}, the authors aimed to address the challenges faced by supply chains (SCs) in make-to-order (MTO) environments, particularly those related to operational resilience and recoverability. Recognizing the limitations of standalone cyber-physical systems (CPS) in controlling SCs under such dynamic conditions, they proposed a coordinated approach through a cyber-physical logistics system (CPLS). This CPLS framework, designed within a multi-level CPS structure, focuses on providing resilient SC control by coordinating distributed DT simulations with agent-based CPS systems. The study further outlines service composition procedures and operation processes within the DT framework to mitigate common SC issues like the bullwhip and ripple effects, demonstrating an early case of effective SC and production planning using DT simulations. Building on the need to address the complexities and vulnerabilities in modern supply chains, the authors in \cite{wang2022digital1} explored how the DT-driven Supply Chain (DTSC) concept can transform traditional supply chains into smart-integrated systems. Recognizing the necessity for a smart supply chain that is connected, visible, and agile, their proposed approach, DTSC, leverages the DT concept to create an integrated and intelligent supply chain.

\subsection{Healthcare and Medicine Industry}

DT technology has significantly advanced the healthcare sector, bringing transformative improvements to various aspects of patient care and data management. By creating real-time, dynamic replicas of physical entities, DTs have revolutionized the management of Electronic Health Records (EHRs) and the personalization of treatment plans via two-way communications between healthcare devices and virtual entities, suchs as via wireless communications over hospital networks or Wifi networks in home settings. This section explores the significant roles of DTs in healthcare, analyzing DT-related communications to support EHRs management and patient treatment. 


\subsubsection{DT for Electronic Health Records (EHRs) Management}
DTs are increasingly being utilized to enhance EHRs management by improving data accuracy, privacy, and interoperability across healthcare systems. For example, in \cite{liu2019novel}, the authors present a new framework integrating DT technology with cloud computing and IIoT to advance elderly healthcare services. This framework, known as CloudDTH, addresses key challenges in personal health management by creating a comprehensive system that bridges physical and virtual healthcare environments. By employing wearable medical devices, CloudDTH facilitates real-time monitoring, diagnosis, and prediction of health conditions. In this framework, CloudDTH utilizes wireless communication to enable seamless data transfer between wearable medical devices and cloud-based systems, supporting real-time health monitoring and predictive analytics. Another study \cite{elayan2021digital} explores a novel DT framework that represents a virtual model of physical assets, reflecting real-time data for improved healthcare management. This framework contributes significantly to digital healthcare by integrating an intelligent, context-aware system with a machine learning-based electrocardiogram (ECG) classifier. The patients communicate through the cellular netwrok and the healthcare professionals receive real-time ECG data from them wirelessly, enhancing the accuracy and efficiency of heart condition diagnosis and management. The ECG classifier effectively diagnoses and detects heart conditions with high accuracy across various algorithms. 

Moreover, the challenge of safeguarding data privacy in EHRs is addressed through an advanced privacy-preserving framework discussed in \cite{garg2022spoofing}. This framework integrates DT technology with Cyber-Physical Systems (CPS) to streamline the collection and evaluation of health metrics from various sources, such as wearable devices and medical instruments. This approach creates a virtual counterpart of physical health data, enabling real-time monitoring via wireless networks and improved diagnostic capabilities. The proposed solution features a two-phase framework using EfficientNet Convolutional NNs to effectively differentiate between authentic and spoofed iris samples. According to \cite{coorey2022health}, the application of DTs in CVD is a rapidly developing field with promising potential.     Despite the promising advancements, the implementation of DTs faces challenges such as ethical considerations and clinical integration barriers. These hurdles highlight the need for ongoing research and development to fully realize the benefits of DTs in revolutionizing cardiovascular care.
\subsubsection{DT for Personalized Treatment}
DTs are emerging as a transformative tool in the field of precision medicine, offering significant advancements in the creation and integration of computational models for individual patient care. For instance, this study \cite{masison2021modular}  introduces a modular software platform designed to facilitate the development of medical DTs, which are computational models of disease processes tailored to individual patients using diverse data sources. The proposed framework mirrors industrial strategies used in preventive maintenance by integrating mechanistic and data-driven approaches to model various medical conditions. The platform leverages advanced communication protocols to enable efficient data exchange and collaborative development across research labs, supporting the creation and updating of personalized medical DTs. The authors in \cite{erol2020digital} delve into how DTs are employed to create patient-specific models, enhancing clinical decision-making and treatment effectiveness. The study highlights how DTs use robust communication systems to synchronize patient data and medical device information, enabling accurate modeling and personalized treatment plans. By creating detailed digital replicas of patients and medical devices, DT enables precise modeling of individual health conditions and personalized treatment plans.  Recently, Human DT (HDT) technology holds transformative potential for enhancing personalized healthcare services (PHS) is explored in \cite{okegbile2022human}. Analogous to DTs in sectors such as manufacturing and aviation, HDT comprises three integral components: the physical entity, the virtual model, and the interactive linkage between them. HDT presents more intricate challenges compared to its industrial counterparts, with its implementation strategies still being refined. The HDT system communicates wirelessly with wearable sensors and medical devices to continuously update the virtual model with real-time physiological data.


\subsection{Transportation and Logistics}

\subsubsection{DT for Fleet Management}
DT technology enhances fleet management by offering a dynamic and accurate representation of vehicles and their operations. This advanced technology enables fleet managers to monitor every vehicle's performance in real time, predicting maintenance needs, optimizing routes, and improving overall efficiency without the need for physical inspections. For example, in \cite{alexandru2022digital}, DTs are employed to foster collaboration, improve factory layout modeling, and enhance Automated Guided Vehicles (AGVs) control software. The DT system communicates with AGVs via the cellular network and control software to optimize fleet scheduling and route planning, enhancing operational efficiency and reducing delays in high-traffic environments like airports. It describes a prototype Industry 4.0 factory featuring multiple AGVs and details an ongoing project to develop DT for these vehicles. Optimizing specialized vehicle scheduling is crucial for efficient airport operations, particularly during high traffic to reduce delays and improve passenger satisfaction. Recently a collaborative scheduling model for different vehicle types aimed at minimizing travel distance and waiting time was proposed in \cite{luo2024multi}. It introduces a three-layer genetic algorithm with advanced crossover and mutation techniques to address the scheduling complexities. Given the unpredictability in airport operations, the paper combines this genetic algorithm with simulation methods and DT  technology to create a multi-strategy scheduling framework. The multi-strategy scheduling framework communicates real-time data through DTs to dynamically adjust vehicle schedules, enabling rapid re-scheduling in response to operational delays.

Meanwhile, the authors investigated another significant sector, Maritime Transportation Systems (MTS) in \cite{liu2021security}. Due to the current limitations in MTS security, the study develops a DT model that incorporates relay nodes to enhance data transmission in MTS, improving communication security and advancing digitalization in maritime operations.  
Moreover, vehicle health is another key aspect of fleet management. DT enable Integrated Vehicle Health Management (IVHM) in aerospace, which creates a virtual representation of a physical system to simulate and predict its behavior is shown in\cite{ezhilarasu2019understanding}. IVHM aims to enhance Condition-Based Maintenance (CBM) by continuously monitoring and analyzing the health of vehicles. DTs in IVHM communicate real-time health data from physical systems to virtual models, enabling continuous monitoring and predictive maintenance for complex systems like aircraft.
However, implementing DT technology in fleet management presents challenges. DTs and Cyber Physical Systems (CPSs) hold significant promise for enhancing the intelligence and efficiency of commercial vehicles like buses and trucks. The authors in \cite{de2024digital} addresses these issues, offering solutions and strategies for overcoming technical and organizational hurdles in the adoption of DT. Key issues include a lack of consensus on the definitions of DT and CPS and a tendency to focus on only a single dimension of physical assets, which limits their comprehensive use.

\subsubsection{DT for Traffic Management}
The significance of DT technology in traffic management is underscored by its ability to provide a dynamic virtual model of real-world traffic conditions. In the context of the Internet of Vehicles (IoV), the vast amounts of traffic data pose significant challenges to effective traffic resource scheduling. To overcome these challenges, the authors in \cite{hu2021digital1} introduce a short-term traffic flow and speed prediction method called TFVPtime-LSH. This method communicates traffic data captured by distributed cameras through 5G networks to enhance short-term traffic flow and speed predictions. The effectiveness of TFVPtime-LSH in forecasting short-term traffic conditions is demonstrated through experiments conducted on real-world traffic data from Nanjing, China.
Moreover, a recent study \cite{liao2024digital} introduces a Social Value Orientation (SVO)-based cooperation mechanism for AVs, which determines driving routes based on individual needs, local road conditions, and overall benefits. The DT-based edge-to-cloud traffic guidance architecture communicates real-time AV decisions and micro-driving data to estimate future road conditions, optimizing route planning while reducing communication and computation overheads. This hierarchical structure effectively reduces communication and computation overheads by distributing tasks across different levels. Another important sector is the Airport Traffic Control system, where DT technology is being utilized to enhance centralized traffic control solutions. Some experimental work has explored the potential of DT in this domain; for example, in \cite{saifutdinov2020digital}, the authors present experiments with a DT specifically designed as a testbed for such applications. The DT system communicates simulated data on vehicle spatial characteristics and critical scenarios to enhance centralized traffic control and manage complex situations within the transport network. Furthermore, significant advancement in smart city infrastructure is highlighted in \cite{xu2023smart}, which introduces the design, implementation, and use cases of the Chattanooga DT (CTwin) aimed at revolutionizing next-generation smart city applications for urban mobility management. The CTwin platform communicates multi-domain urban mobility data from online sources and IIoT sensors to provide a comprehensive view of traffic, hazards, weather, and safety for advanced smart city management.
Moreover, as physical networks continue to see increased bandwidth and faster speeds, a comprehensive flow emulation framework has been developed for DT networks \cite{yang2021systematic}. This framework uses unified ID and deterministic network technology to keep traffic consistent between the physical network and its DT. Additionally, it incorporates flow sampling to manage the increasing data volume.

\subsection{Energy Industry}
In recent years, the energy industry has faced growing challenges, including the need for more efficient resource management, enhanced system reliability, and the integration of renewable energy sources. By leveraging DT technology, energy companies can simulate various scenarios, predict system behaviors under different conditions, and optimize operations to reduce energy waste and costs


\subsubsection{DT for Grid Management and Optimization}
 A range of DT-based systems has been developed to improve grid management and optimization, which are critical in contemporary energy systems for efficient power distribution and maintaining grid stability. For example, an innovative method for identifying time-varying load dynamics is proposed in \cite{baboli2020measurement}. This approach combines system identification techniques with nonlinear numerical optimization, leveraging artificial neural networks (ANNs) to link model parameters with real-world measurement data. Connectivity is established by linking ML-based servers with smart grid sensors over Wi-Fi, ensuring secure and reliable real-time data analysis and optimization. While this ANN-based method improves real-time system identification, challenges remain, particularly in accurately modeling reactive power dynamics under varying conditions.
 \begin{figure}[h]
     \centering
     \includegraphics[width=\linewidth]{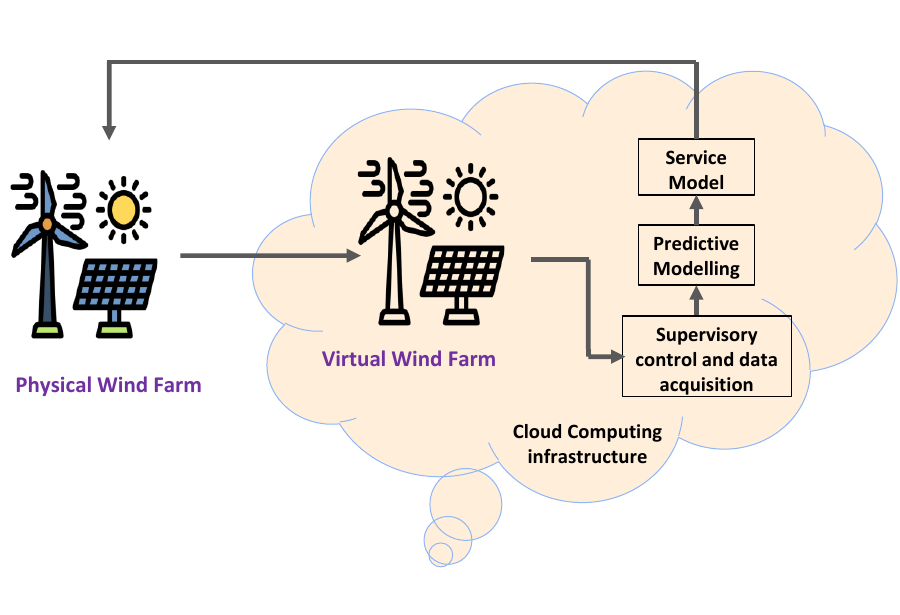}
     \caption{  A cloud-based framework that utilizes DT to offer bidirectional communications between physical and virtual wind farms for power generation management.}
     \label{Wind farm}   
\end{figure}
Fog computing is also increasingly recognized for its role in reducing latency and enhancing performance by storing information in the cloud, which enables the delivery of diverse services. A study in \cite{fan2023energy} proposes a three-layer model comprising cloud, fog, and user layers to optimize energy management in renewable power grids, utilizing DT technology for precise monitoring. An innovative Whale Optimization Algorithm (WOA) is introduced for effective fog load balancing. By integrating DT with WOA, this approach offers a promising solution for managing renewable energy resources. The use of Fog servers and IIoT networks, connected through wireless connections, enhances real-time data processing and decision-making, potentially reducing costs and boosting energy efficiency in the power grid.
Another study in \cite{gao2023stochastic} proposes a distributed Energy Management System (EMS) that separates the optimization process into two tiers: one for the overall microgrid and one for local controllers, considering both the network constraints and load uncertainties. The Moth Flame Optimization Algorithm (MFOA) is used to solve this problem within a DT framework, which includes solar panels, wind turbines, diesel generators, and batteries. Moreover, this study \cite{dellaly2024digital}  highlights that adopting renewable energy with this peer-to-peer trading model can enhance energy self-sufficiency and lower reliance on external sources, leading to financial savings. The proposed EMS, tested on a commercial microgrid platform, shows promising results. Promoting the continued evolution of DT applications recently, a DT-based platform for managing DERs that leverages cutting-edge Industry 4.0 technologies are introduced \cite{kovalyov2022distributed}. The platform aims to improve power supply quality, cut costs, and explore new market opportunities. It is built using a viewpoint-based framework, as defined by ISO/IEC/IEEE standards, and integrates a typical DT architecture for power systems. 

\subsubsection{DT for Energy Storage and Renewable Integration}
DTs offer substantial benefits for optimizing energy storage and integrating renewable sources into power systems. For example, the integration of a physics-based DT for a combustion engine with an electrical power plant model and battery storage is explored in \cite{soderang2022development}. They developed a Fast-Running Model (FRM) from a detailed, crank-angle resolved engine model and paired it with a power plant control model created in Simulink. Leveraging Ethernet for real-time communication between the simulation and physical layers, the integrated engine-electric model demonstrated a slight decrease in accuracy compared to simpler simulations, but it was still sufficiently accurate for control and monitoring tasks.
To address the challenges forecasting power generation and managing wind farms due to the unpredictability of wind speed, a cloud-based DTs architecture is proposed in \cite{fahim2022machine}, as shown in Fig. \ref{Wind farm}. 
This model is based on the DT infrastructure of Microsoft Azure and makes precise predictions by analyzing time series data using sophisticated deep learning methods. Wind speed sensors transmit data to Microsoft Azure via cellular networks and Wi-Fi, enabling real-time analysis and accurate forecasting.
To improve the stability and efficiency of energy management for a large wind farm, a multi-functional Battery Energy Storage System (BESS) is implemented in \cite{li2024multi}. DT technology has been instrumental in fine-tuning the BESS capacity, ensuring it meets both stability and economic goals. The findings confirm that integrating multiple control strategies can be done without adverse effects, marking a significant step toward the practical implementation of such a versatile BESS. Another approach provides a dynamic forecast of battery behavior, which helps in managing the system more effectively and safeguarding against faults and cyber threats \cite{kharlamova2022digital}. The study utilizes real-world frequency data from a BESS operating in the Nordic region to test different AI techniques for state of charge (SOC) forecasting. The research finds that data-driven models, particularly those using ANNs, offer reliable and precise SOC predictions. 

Furthermore, DT-based secure thermal energy storage in buildings is explored in \cite{lv2023digital}, which can enhance the scheduling of thermal energy storage systems in smart buildings, contributing to more sustainable economic development. This approach allows for simultaneous real-time analysis and optimization. The developed DT-based model integrates phase change material walls and a thermal network to manage energy storage effectively. Also, managing local energy systems becomes increasingly complex, and creating a DT model that combines smart meter data with additional geospatial and building information offers a promising solution \cite{bayer2023digital}. Here DT is designed to simulate and optimize the integration of renewable energy sources such as photovoltaic systems and battery storage, as well as to address the increased energy demands from large appliances like heat pumps. By using this model, system operators, and households can make more informed decisions about where to install renewable energy systems and how to manage energy consumption effectively. 

\subsection{Agriculture and Food Industry}

\subsubsection{DT for Precision Agriculture}
As agriculture evolves due to technological improvements, the incorporation of digital technologies provides new prospects for improving farm management and sustainability. DTs give farmers important operational information that facilitates more effective resource management and improved decision-making by connecting the physical and digital realms. 
Developing a DT model for agriculture, integrating soil probes from the sensing change initiative and the Smart Water Management Platform (SWAMP), is a perfect example \cite{alves2019digital}. By linking agricultural sensors wirelessly to a computing server using mobile networks, the system enhances farmers monitoring and management of resources and equipment.
The system successfully captures data from soil probes and displays it on a dashboard, setting the stage for expanding the network with additional monitoring devices to achieve a fully functional DT. This approach supports better decision-making and helps mitigate the environmental impact on water, soil, and land resources. Similarly, a software-based solution known as the "DT of Rice", which simulates real-time circumstances in rice farms analyzed in \cite{skobelev2021digital}. DT uses an ontology-based knowledge base to apply plant growth principles, allowing for real-time data collection and decision-making for optimizing rice farming. Developed as a standalone service, the DT is designed for seamless integration with existing digital agricultural systems. 
In another work \cite{kim2024agricultural}, a DT structure for agriculture is constructed using mandarins as an example crop. The findings show that, in comparison to a more comprehensive inter-orchard examination, data analysis inside particular orchards yields a more accurate assessment of fruit quality. The authors in \cite{jans2020digital} demonstrated how to optimize energy consumption while enhancing crop growth by managing indoor environmental conditions of a hydrophonic farm in London. Their proposed model can be integrated into the DT to provide actionable feedback for farmers, enhancing decision-making and efficiency. Meanwhile, the author presented the "DIWINE" project in \cite{edemetti2022vineyard}, in which the DT platform transforms the field of smart and sustainable agriculture, particularly for vines, by using unmanned aerial vehicles (UAVs). The platform provides winemakers with immediate and adaptable access to detailed vineyard information and integrates seamlessly with existing technologies like IIoT sensors and weather forecasts. This improves decision-making, lowers the chance of missed harvests, increases profitability, and ultimately results in more informed and efficient vineyard operations.

\subsubsection{DT for Smart Food Production and Supply}

Conventional food supply chain management methods often operate in siloed stages, leading to inefficiencies, slow response to disruptions, and limited scalability. DT technology provides real-time monitoring, predictive insights, and improved coordination throughout all phases, thereby overcoming these constraints. 
For instance, Procurement, Production, and Distribution (PPD) methods in a medium-sized food processing company are optimized through a DT framework, as demonstrated in \cite{maheshwari2023digital}. Utilizing mixed-integer linear programming (MILP) and agent-based simulation (ABS), the model explores the industrial symbiosis between suppliers, manufacturers, and customers within a constrained environment. Real-time data from industrial sensors is transmitted to a cloud server via wireless communication, enabling real-time analysis and optimization of the entire process.
A greater degree of digitization is attained by the DT technique, which also increases makespan, lead time, and overall operations effectiveness. 
Similarly, the concept of "eGastronomic Things" involves creating gastronomic devices that have both physical and digital counterparts, showcasing how DT technology is advancing in the gastronomy sector \cite{karadeniz2019digital}.

Traditionally, preserving the freshness of fruits during refrigerated transport and storage offers major issues due to temperature variations and biochemical degradation. To address these issues, the authors in \cite{defraeye2019digital} developed a DT specifically for mango fruit. 
The study also shows that DTs can be used to show how various cold chain durations and temperature histories affect fruit quality, especially when it comes to perishable fruits like mangoes that are kept in conditions with little ventilation. These insights help identify where and how quality loss occurs, allowing for improvements in refrigeration processes and logistics to reduce food waste. Meanwhile, a different study \cite{vignali2020tube} demonstrated the creation of a DT for a tube-in-tube pasteurizer. Once the model is validated, temperature and pressure sensors on the actual plant and a data acquisition module are used to compare real-time industrial data with the simulated data. This comparison enables operators to monitor the process closely and intervene before any issues arise, thus preventing potential safety problems and product loss. The COVID-19 outbreak exposed serious flaws in typical food retail supply networks. To cope with these problems, the authors in \cite{burgos2021food} investigated how the pandemic affected German supply chains (SC) by combining a discrete-event simulation model with an anyLogistix Digital SC Twin. In response to these findings, the authors proposed several strategies to enhance SC resilience. These include the broader adoption of DTs for real-time monitoring and decision-making, improving end-to-end visibility across the supply chain, and refining demand and inventory management practices to better handle disruptions. 

\subsection{Space Industry}
The space industry is being significantly impacted by recent developments in DT technology. By enabling real-time adjustments based on virtual simulations and predictive maintenance, this technology improves operational efficiency and mission reliability. The application of DTs to spacecraft and satellite management is discussed in this part, with special attention to how they might be used to predict performance, optimize resources, and improve missions overall. 
\subsubsection{DT for Spacecraft Industry}
Maintaining good assembly quality and dependability is crucial as spacecraft get more complicated. Typical computerized models frequently fail because they are overly idealistic and fail to capture the manual, real-time aspect of industrial spaceship assembly. To solve these problems, a novel method of real-time data collecting that seamlessly merges with the assembly procedure is presented in \cite{yang2022digital}. 
The authors developed a DT system for spacecraft assembly, updating the virtual model with real-time data from a sensor network connected to cloud computing, enabling rapid feedback and problem-solving. This method offers an effective way to raise assembly quality and clarify the production process in spaceship assembly by contrasting real-time monitoring with numerical analysis. In exploring the application of digital technology to spacecraft component design, this article \cite{khodenkov2023digital} emphasizes the concept of \lq digital space\rq\ as a basis for improving the design process. With the rising complexity of aerospace systems and increasing demands for quality and dependability, the study investigates how digital models might streamline various stages of a product's lifespan. The paper specifically examines the use of DTs to evaluate the mechanical properties of a carbon fiber rod used in a deployable reflector. This approach not only speeds up the design process but also helps reduce costs by replacing physical tests with virtual simulations. 
Conventional techniques, such as laser-driven flyer (LDF) experiments, have limitations because of their expensive and hard-to-control conditions, which reduces the amount of data that can be used to assess damage. The author in \cite{wang2022digital3} tried to find a solution by presenting a new method based on LDF DTning, which consists of two main stages. During the experimental phase, the approach ensures a standardized environment and allows for automated and remote experiments, enhancing both efficiency and the ability to conduct virtual training. While in the evaluation phase, NNs are used to expand limited experimental data into comprehensive datasets with detailed damage parameters. Furthermore, finite element simulations provide information about microscopic damage processes. Results show that this method achieves an average error rate of 8.13\% for neural networks and 3.1\% for simulations, both of which are below the 10\% error barrier set by the standard method.

\begin{figure*}[h]
  \centering
      \includegraphics[width=\linewidth]{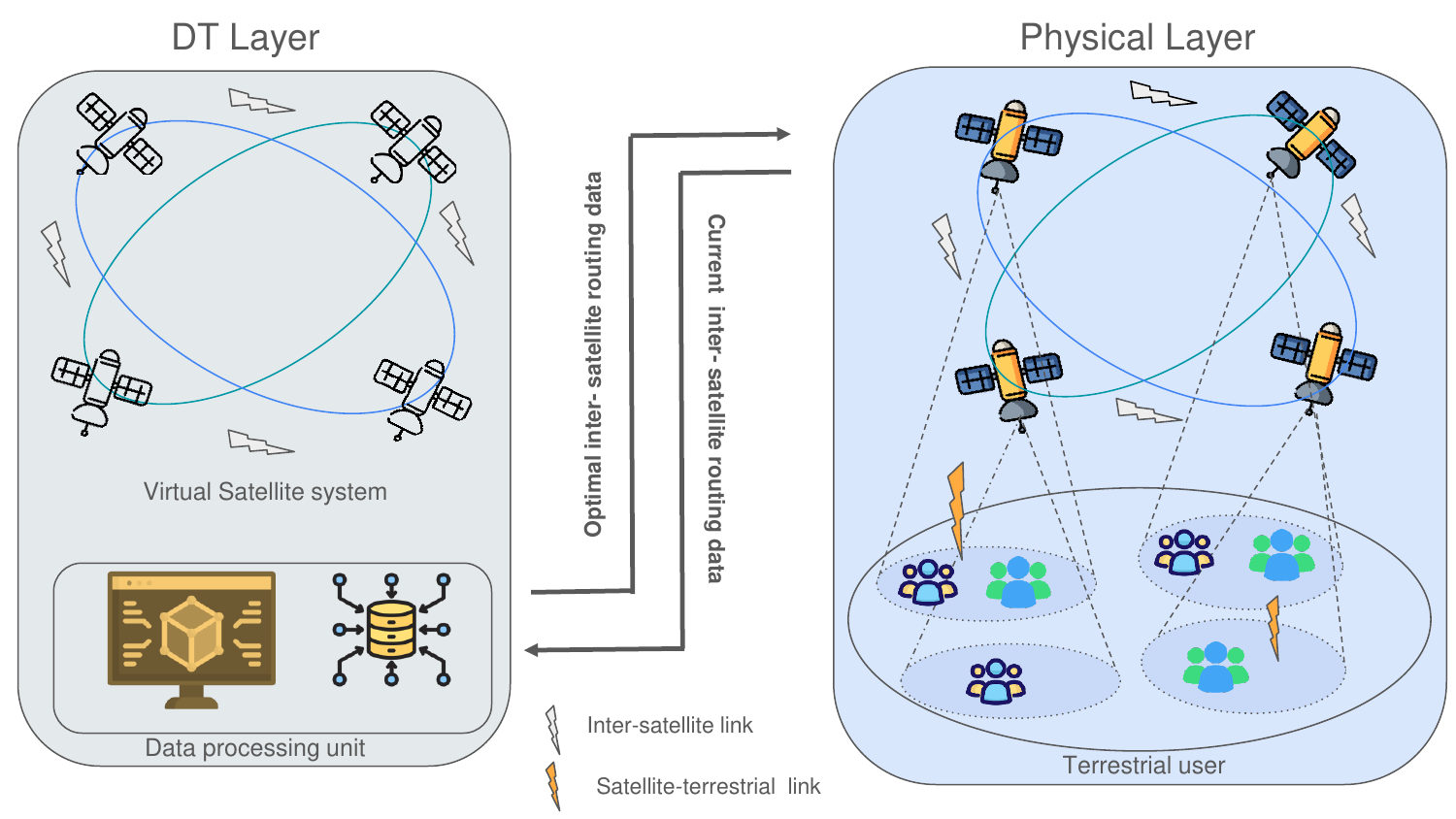}
      \caption{ System model illustrating the digital twin-assisted inter-satellite routing scheme, enhancing data delivery quality by considering satellite link timing visibility.} 
      \label{sattelite}   
\end{figure*}
Astronaut training, which requires costly and specialized equipment, is being enhanced by integrating immersive technologies with DTs. The authors in \cite{pinal2024mixed}, developed four scenarios across three modules based on NASA and ESA procedures. The first module uses mixed reality (MR) for theoretical training. The second module provides realistic simulations with DTs of ISS systems and a spacesuit, including emergency scenarios and spacewalk repairs. The third module replicates a SpaceX Falcon Heavy launch using DTs and 3D models of the rocket’s systems. Interactions with the simulations are compared to actual data from Falcon Heavy launches and the ISS. Connectivity is established by linking the computing server with ISS propulsion and navigation systems and spacesuit sensors for real-time data integration and analysis. The results showed how successfully DTs complemented immersive technology by demonstrating an imitation of the real systems and proved to be beneficial astronaut training tools. Meanwhile, a DT framework for managing and tracking the health of these spacecraft is presented in \cite{ye2020digital}. Reusable spacecraft are transforming space travel by lowering costs, but maintaining structural integrity between trips is vital. The suggested framework provides features like performance evaluations, model updates, and diagnosis in two phases: offline and online. They simulate the emergence of fatigue cracks in a load-bearing frame to show the capabilities of the framework. The model adjusts dynamically as cracks form, increasing the precision of future predictions. This ongoing improvement makes evaluating the spacecraft's reusability more accurate. 

\subsubsection{DT for Satellite Industry}
Assembling small satellites in orbit requires precision and adaptability, which is challenging due to the difficulty of automating satellite manufacture in space. To address these challenges, the DT-based ‘AI In Orbit Factory’ project proposes an automated production system specifically for space-based satellite assembly \cite{leutert2024ai}. This system uses AI and ML-based servers to identify issues in optical and electrical systems early and prevent mission-threatening mistakes. It integrates an industrial IIoT network, where sensors share data via Wifi, allowing for precise management of tolerances and fluctuations. Force-sensitive measurement techniques ensure components are assembled with the required precision for orbit. The system also features AI-guided teleoperated control, providing additional flexibility and durability by allowing human operators to intervene when necessary. Results are promising, showing that this approach significantly improves the accuracy and reliability of satellite production in space. By detecting and correcting potential issues early, the system reduces the risk of deploying faulty satellites, which is crucial for the success of space missions. Furthermore, the harsh conditions in space make it difficult to diagnose problems and keep track of the health of increasingly complex satellite systems. Conventional techniques for fault diagnosis and health monitoring (FD-HM) frequently rely on physical data that is static and historical, which limits their capacity to offer real-time maintenance and insights. A DT-driven method for FD-HM is suggested in \cite{shangguan2020digital} to overcome these drawbacks. With this method, real-time telemetry, fused data, and simulation data are integrated to create a DT that combines virtual and physical models. Real-time monitoring and maintenance are made possible by the DT's sophisticated data-driven and model-based algorithms, which offer a dynamic and thorough picture of the satellite's operational state. A practical demonstration is also offered by a DT-based space ground power management platform, which enables real-time visualization and monitoring of the satellite power system with minimal manual effort. This strategy represents a substantial development in satellite operation and maintenance technology and has the potential to spread its advantages to additional satellite subsystems, potentially even entire satellite systems.

Gaining attenuation in Low-Orbit satellite (LEO) networks is another key concern, despite their extensive coverage and low latency. Satellites travel faster than ground terminals, which causes frequent handovers and inconsistent service. INTERLINK, a DT \- assisted approach to improve satellite-terrestrial network performance, is introduced in \cite{zhao2022interlink} as a solution to these issues, shown in Fig. \ref{sattelite}. Initially, a new handover strategy named ASHER is considered to reduce the frequency of handovers by taking into account the satellites' remaining service periods, capacity limitations, and restricted access hours. Using a genetic algorithm, this technique solves the challenge by framing it as a multiobjective optimization problem and identifies the best satellites for handovers. Thus the INTERLINK scheme considers the dynamic nature of the satellite network and optimizes routing paths using a time-varying graph. Based on simulation results, INTERLINK improves the quality of data delivery while achieving a significant reduction in handover frequency and average propagation delay while increasing routing efficiency. Advanced satellite systems with higher bandwidth are required due to the growth of mobile phone services, particularly for in-flight connectivity, e-commerce, and widespread internet use. The author in \cite{zohdi2024machine} suggests using a DT-based "mega-constellations" of satellites, combining different kinds in low, medium, and geostationary orbits, to satisfy this need. These constellations use a wide range of frequencies to allow continuous, high-bandwidth, low-latency communication, something that is not possible for single-orbit satellite systems. Specifically, the DT framework is intended to efficiently simulate and optimize constellations of satellites around a fictional planet known as \lq Planet-X.\rq\ To get the best coverage for communication or imaging needs while taking resource limits into account, this framework makes it easier to explore various satellite infrastructure configurations, including the number, orbits, speeds, and types of satellites. Moreover, DT can be used for runtime verification to protect satellite systems from cyber attacks. This paper \cite{hou2022digital} illustrates a technique for creating and coordinating DTs to guarantee reliable and secure communication between satellites and their ground equivalents, taking into account the particular requirements of space missions. The framework features a runtime verification engine capable of assessing properties across various temporal logic languages. Future plans include penetration testing and attack simulations to gather data on system operations and security. This data will inform Markov decision processes and reinforcement learning for optimizing strategies. The Process Analysis Toolkit (PAT) will be used for formal verification of satellite systems and synchronization protocols, employing Petri nets for modeling and validation.
Results showed a completely verified execution stack that guarantees secure communication and functionality for satellites and other space assets.

\subsection{Oil and Gas Industry}
The management of assets, operating effectiveness, and safety have always been difficult issues for the oil and gas sector, especially in hostile and isolated locations \cite{mojarad2018challenges}. Companies frequently depended on reactive maintenance and sporadic inspections in the absence of sophisticated monitoring and predictive systems, which raised operating costs, created unanticipated downtime, and raised the risk of safety issues. 
DT significantly reduces the shortcomings of traditional methods and enhances industry-wide decision-making, safety, and efficiency by enabling continuous monitoring, predictive maintenance, and optimization of operations \cite{wanasinghe2020digital}. From several key domains where DT technology benefits the oil and gas industry, we discussed two: project management and safety assessment, and production optimization.

\subsubsection{DT for Project  Management and Safety Assessment}
The oil and gas sector has seen a revolution due to the integration of digital assets, which has made it possible to solve the intricate problems associated with deepwater operations, high-pressure, high-temperature wells, and sophisticated algorithms. Operators must make better decisions as they deal with more and more difficult oilfield management issues. DT technology enables real-time monitoring of wells, offering insights into construction, costs, and project-specific issues. Major oil operator oversees the management of over 200 wells each year, including those with significant depths over 20,000 feet and extensive lateral intervals exceeding 8,000 feet. To navigate these complex and costly challenges, the company employs DTs to monitor real-time well performance, covering aspects such as drilling, completion, and production \cite{lai2022digital}. 
This method, combined with industrial big data analytics, enhances decision-making efficiency. Sensors share data wirelessly with the computing server, addressing challenges like optimizing well construction and evaluating return on investment. Consequently, the company can manage operations more effectively, with reduced manpower and increased capacity. Upstream employees in the oil and gas industry have long suffered from ineffective data gathering and analysis; they frequently spend as much as 80\% of their time locating and converting data from several sources. This inefficiency, which results from fragmented data systems, costs more and wastes a lot of time. The authors in \cite{brewer2019digital} introduce “FieldTwin”, a DT-based FutureOn product, which can overcome these obstacles. FieldTwin enables seamless communication by connecting IIoT sensor networks to cloud computing, breaking down data silos, and making real-time data accessible across the organization.. This allows employees to spend less time on data retrieval and more on analyzing and utilizing data, which can enhance drilling strategies, field automation, and safety measures. Besides, FieldTwin provides visual representations of field data in real-time, which enhances decision-making and data understanding. Project management and risk assessment are made easier with FieldTwin's visual tools since people assimilate visual information far more quickly and efficiently than text.
In \cite{correa2023process}, the author investigates a cloud-based platform that uses DT technology to increase safety and operational efficiency in the oil and gas industry. By combining project data and process documentation into a single, centralized DT database, the platform provides industrial operations with a comprehensive tool for risk management and safety oversight.
Through web and mobile applications, operators can utilize the platform to access data and models and run simulations. The solution, which was created especially for Floating Production Storage \& Offloading (FPSO) units, links risk analysis data to a 3D model by extracting and contextualizing the data using regular expressions and OCR. Then it integrates middleware for handling 3D models and a document integrator for risk analysis, providing dynamic visualizations that illustrate the sequence of actions needed during process interventions. By centralizing engineering information, the system improves operational efficiency and reduces the time needed for data retrieval and validation by up to 75\%. Additionally, it increases safety by providing illustrations and simulations that help with intervention planning and decision-making.

Similarly, a strategy for predicting possible pipeline system breakdowns by combining prognostic algorithms and ML is presented in this study \cite{priyanka2022digital}. They used prognostic models to assess risk based on pressure data, as well as the use of clustering techniques like Dirichlet Process Clustering and Canopy Clustering to detect aberrant pressure fluctuations. Data from multiple oil substations are analyzed through manifold learning to extract relevant features, which are then assessed using a kernel-based Support Vector Machine (SVM) to predict risk probabilities. By providing an effective wireless data connection between the server and oil substations, the framework makes real-time control possible. This led to the development of a virtual intelligent integrated automated control system, which uses advanced wireless technology to improve risk assessment and management throughout pipeline networks, particularly in distant areas.  The "Tree of Consequences" hazard analysis technique is an alternate solution that is commonly employed in oil and gas facilities to ensure industrial safety is explored in \cite{abdrakhmanova2020review}. It gives an overview of software solutions that improve risk analysis using 3D modeling and visualization. For example, in terrestrial settings, FLACS is used to model ignition and toxic emission scenarios. The effects of hazardous materials on manufacturing facilities are evaluated by the TOXI-Risk software, and the complete chain of events leading up to the eventual damage is simulated using Phast Lite. DT facilitates constant communication and data exchange with physical objects. This technology holds significant promise for enhancing maintenance, preventing accidents, and managing production.

\subsubsection{DT for Production Optimization}
DT technology, bolstered by advancements in IIoT and AI, significantly increases the efficiency of oil and gas industry production throughout the asset lifespan. For example, this study \cite{shen2021digital} discusses the deployment of DTning in the oil and gas industry, presenting a model that incorporates entities, models, data collecting, intelligent algorithms, service, and interface control. 
It allows for real-time data integration by wirelessly linking the oilfield sensor network to an NN server, resulting in a 3\% increase in efficiency, improved design accuracy, and good operational performance. 
Recently, a study \cite{latif2019holistic} examined various carbonate reservoirs in an aging gas field with over 30 years of production history and more than 150 gas condensate wells. To address the inevitable production decline, the study evaluated different approaches to meet Service Level Agreements (SLAs), extend the production plateau, and optimize operating costs. Connectivity is achieved by integrating the computing server with an IIoT sensor network, facilitating detailed data analysis and enhancing operational efficiency. Reducing backpressure in surface facilities is a prominent subject of study in oil and gas fields since it has a significant impact on production rates. Among the strategies explored, reducing the inlet and outlet pressures at the gas plant was found to be highly effective in extending the production plateau. Field tests were used to validate this strategy and show how workable it is. Promising results were found in both simulation and subsequent pilot implementations when the study looked into the benefits of wellhead compression for wells with poor or sporadic output. An important tool for comparing different gas field management approaches was the Integrated Asset Model (IAM). It functioned as a DT, enabling the modeling of situations like the addition of new infill wells to increase production—a move that needed rigorous economic analysis. Meanwhile, the author presents a new method of maximizing the use of natural gas by utilizing a combination cooling, heating, and power (CCHP-CER) system, which effectively captures the thermal and cold energy of Liquified Natural Gas (LNG) in \cite{huang2022digital}. The system combines four essential industrial parts that are all controlled by a DT framework: a heat exchanger, a cold energy recovery unit, an absorption chiller, and a gas turbine. This DT framework, developed using a cascade forward neural network (CFNN), allows for real-time modifications and life-cycle optimization of the system's operational parameters. Notably, the addition of the cold energy recovery unit not only improves electricity production and cooling capacity but also results in a 0.72\% rise in the daily Primary Energy Savings Rate (PESR). Moreover, the optimization process driven by DTs leads to a significant improvement in energy efficiency, especially during times of system degradation. This process achieves higher energy savings than traditional static models, with gains of 2.23\% in winter, 0.35\ in summer, and 1.53\% during transitional seasons. In contrast to conventional approaches for the petrochemical industry, this study \cite{min2019machine} presents a novel theoretical framework for production control based on DT technology. The suggested approach eliminates the need for expert knowledge and the limits of single machine learning outcomes. Rather, it uses industrial big data to continuously develop and refine dynamic models that respond to changes in the environment. This method addresses key challenges in big data analysis and model training within the petrochemical industry. This approach is not only applicable to the petrochemical sector, but it also provides an innovative approach for enhancing economic performance in a variety of process manufacturing businesses through improved production control. The study addresses how simulation and dynamic modeling have developed into valuable tools for increasing upstream operations' profitability. Regardless of their promise, these approaches have hurdles, such as the necessity for independent models and workflows across different sectors, including wells, gas-oil separator facilities, and electrical submersible pumps. A DT-based solution presented in \cite{okhuijsen2019real} addresses these problems by using synchronous equation-based modeling, providing a more accurate and dependable approach to closed-loop production optimization. With this method, process control systems can instantly apply optimum setpoints, eliminating the need for manual checks and propelling the sector toward greater automation.
\begin{table*}[ht!]
\centering
\caption{Taxonomy of DT applications in industries.}
\begin{adjustbox}{max width=\textwidth}
\label{tab:DT_applications}
\renewcommand{\arraystretch}{1.5} 
\fontsize{100pt}{120pt}\selectfont 
\begin{tabular}{|c|c|c|c|c|c|c|}
\hline
\textbf{Industry} & \textbf{Use case} & \textbf{Ref.} & \textbf{Digital Platform} & \textbf{Physical Platform} & \textbf{Key contribution} & \textbf{Limitations} \\ \hline


\multirow{6}{*}{\rotatebox{90}{\makecell{Manufacturing Industry}}}
    & \makecell{Product Design\\and Development} & \cite{tao2019digital} & Computing server & \makecell{IIoT network} & A DTPD method for product design & \makecell{Virtual-physical integration\\has not been fully addressed} \\ \cline{2-7} 
    & \makecell{Product Design\\and Development} & \cite{lin2021evolutionary} & ML-based servers & \makecell{IIoT network} & \makecell{An EDT framwork for\\industrial product development} & lacks real-time adaptability \\ \cline{2-7} 
    & \makecell{Product Design\\and Development} & \cite{howard2019digital} & Computing server & \makecell{Industrial IIoT\\network} & A scheme for design optimization & Difficulties in real-world validation \\ \cline{2-7}
    
    & \makecell{Supply Chain\\Management} & \cite{marmolejo2020design} &\makecell{ Cloud computing \&\\computer server}& \makecell{RFID and\\IoT network} & \makecell{A Multi-layered DT architecture\\ to optimize\ the present and\\ even predict the future performance.} & Integration complexity \\ \cline{2-7} 
    & \makecell{Supply Chain\\Management} & \cite{park2021architectural} & Cloud computing & \makecell{Cyber-physical\\systems} & \makecell{A multi-level CPLS architecture \\integrating distributed DT simulation\\to enhance supply chain\\resilience and control} & \makecell{Complexity in multi-level\\coordination and limited\\case study validation} \\ \cline{2-7} 
    & \makecell{Supply Chain\\Management} & \cite{wang2022digital1} & Cloud computing & Sensor network & \makecell{An integrated DT-driven\\approach for enhanced supply\\chain visibility and agility} &  \makecell{The scalability with more manufacturing\\industry has not been investigated.}\\ 
   \cline{2-7} 
    \hline

\multirow{6}{*}{\rotatebox{90}{\makecell{Healthcare and \\ Medicine Industry}}}
    & EHRs Management & \cite{liu2019novel} & Cloud computing & \makecell{Wearable sensor\\network} & \makecell{A CloudDTH framework for\\personalized elderly healthcare,\\integrating physical and\\ virtual medical spaces} & \makecell{Scalability of CloudDTH for varied\\healthcare settings and larger\\patient groups remains untested.} \\ \cline{2-7} 
    & EHRs Management & \cite{elayan2021digital} & ML-based servers & \makecell{Healthcare\\IoT network} & \makecell{A DT-based context-aware\\healthcare system for real-time\\heart condition monitoring and prediction} & \makecell{The scalability with more\\diverse heart conditions and healthcare\\metrics has not been investigated.}\\ \cline{2-7} 
    & EHRs Management & \cite{coorey2022health} & Cloud computing & \makecell{Image sensor\\ network} & \makecell{A DT-based scheme for CVD\\to enhance personalized treatment\\and diagnosis through real-time\\data integration} & \makecell{Ethical challenges and clinical barriers\\to the adoption of AI-driven decision\\tools in real-world healthcare settings.} \\ \cline{2-7}
    
    & \makecell{Personalized\\Treatment} & \cite{erol2020digital} & Computing server& \makecell{Medical device\\sensor network} & \makecell{A framework for leveraging patient DTs\\to improve diagnosis, treatment, and\\ personalized medicine.} & \makecell{Limited focus on practical implementation\\challenges and integration with existing\\healthcare systems and processes.} \\ \cline{2-7} 
    & \makecell{Personalized\\Treatment} & \cite{okegbile2022human} & \makecell{AI and Blockchain\\based cloud computing} &\makecell{IoT and\\mobile netwrork} & \makecell{An architectural framework\\for HDTs emphasizing personalized\\healthcare services through\\AI and blockchain integration}. &\makecell{The implementation methods for HDT\\remain unclear, requiring further\\exploration and validation.}\\ \cline{2-7} 
    \hline
\multirow{6}{*}{\rotatebox{90}{\makecell{Transportation and\\ Logistics}}}
    & Fleet Management & \cite{alexandru2022digital} & \makecell{AI \& ML-based\\servers} & Sensor network & \makecell{A Multi-agent DT environment\\integrating AI for predictive maintenance\\and real-time optimization.} & \makecell{Limited adaptability to \\diverse manufacturing environment\\and operational scenarios.}\\ \cline{2-7} 
    & Fleet Management & \cite{luo2024multi} & Cloud computing & \makecell{IIoT network} & \makecell{A DT-integrated scheduling framework\\ with enhanced genetic algorithms for\\efficient airport vehicle coordination.} & \makecell{The framework's performance\\in more complex airport\\scenarios remains unexplored.}\\ \cline{2-7} 
    & Fleet Management & \cite{liu2021security} & Cloud computing & \makecell{Maritime IIoT\\network} & \makecell{A maritime transportation DT model\\enhancing communication and security\\performance through relay cooperation IIoT.} & \makecell{The model's scalability across different\\maritime routes and varying environmental\\conditions has not been fully explored.} \\ \cline{2-7}
    
    & Traffic Management & \cite{hu2021digital1} & \makecell{5G enabled\\computing server} & \makecell{IoV sensor\\ network} & \makecell{A DT-assisted method for predicting\\real time traffic flow and velocity\\using 5G-enabled IoV data.} & \makecell{Sensor failures and data\\sparsity effects on prediction accuracy\\have not been fully assessed.} \\ \cline{2-7} 
    & Traffic Management & \cite{liao2024digital} & Edge computing & \makecell{AV sensor\\network} & \makecell{A social Value orientation-based\\traffic guidance and Edge-to-Cloud\\architecture for AVs using real-time data.} & \makecell{The impact on traffic guidance accuracy\\with varying AV densities and complex\\urban settings is not fully evaluated.} \\ \cline{2-7} 
    & Traffic Management & \cite{xu2023smart} & Cloud computing & \makecell{Traffic sensors\\network} & \makecell{A CTwin platform for optimizing\\urban mobility with real-time\\analytics and cyber-physical control.} &  \makecell{CTwin's adaptability to unexpected\\urban mobility changes and real-time\\event handling needs further evaluation.}\\ \cline{2-7} 
    \hline
\multirow{6}{*}{\rotatebox{90}{Energy Industry}}
    & \makecell{Grid Management\\and Optimization} & \cite{baboli2020measurement} & ML-based servers & \makecell{Smart grid\\sensors} & \makecell{A method integrating system\\identification with neural networks\\to optimize DER utilization by\\identifying time-varying load dynamics.} & \makecell{The model's validation across diverse\\grid conditions and weather\\patterns remains incomplete.}\\ \cline{2-7} 
    & \makecell{Grid Management\\and Optimization} & \cite{fan2023energy} & Fog servers & \makecell{Industrial IIoT\\network} & \makecell{A three-layer fog computing\\model using WOA for optimized\\energy management and load\\balancing in renewable grids.} & \makecell{The model's performance in grids with\\ varying configurations and its ability to\\handle different renewable energy profiles\\have not been fully tested.}\\ \cline{2-7} 
    
    & \makecell{Energy Storage and\\Renewable Integration} & \cite{soderang2022development} & Computing server & \makecell{Cyber-physical\\systems}& \makecell{A real-time, physics-based engine\\model (FRM) developed for simulation and\\ optimization of power plant operations.} & \makecell{The FRM's execution time needs\\optimization for effective hardware\\in-the-loop implementation.} \\ \cline{2-7} 
    & \makecell{Energy Storage and\\Renewable Integration} & \cite{fahim2022machine} & Microsoft Azure & \makecell{Wind speed\\measurement sensors} & \makecell{A cloud-based DTs framework\\with 5G-NG-RAN for enhanced wind\\speed and power generation\\prediction using deep learning.} & \makecell{Accuracy in extreme wind conditions\\and scalability to diverse wind farms\\are not fully validated.} \\ \cline{2-7} 
     \hline

\end{tabular}
\end{adjustbox}
\label{tab:dt_applications}
\end{table*}

\begin{table*}[ht!]
\caption{Taxonomy of DT applications in industries (continued).}
\begin{adjustbox}{max width=\textwidth}
\label{tab:DT_applications_continued}
\renewcommand{\arraystretch}{1.5} 
\fontsize{100pt}{120pt}\selectfont 
\begin{tabular}{|c|c|c|c|c|c|c|}
\hline
\textbf{Industry} & \textbf{Use case} & \textbf{Ref.} & \textbf{Digital Platform} & \textbf{Physical Platform} & \textbf{Key contribution} & \textbf{Limitations} \\ \hline


\multirow{6}{*}{\rotatebox{90}{\makecell{ Agriculture and\\Food Industry}}}
    & \makecell{Precision Agriculture} & \cite{alves2019digital} & Computing server & \makecell{Agricultural monitoring\\sensor network} & \makecell{An agricultural DTs scheme\\using IIoT and CPS from\\the SWAMP project for\\improved resource management.} & \makecell{Effectiveness with multiple probes\\and diverse farm settings\\has not been investigated.}\\ \cline{2-7} 
    & \makecell{Precision Agriculture} & \cite{skobelev2021digital} & computing server & Field sensor network & \makecell{A platform "DT of rice" for\\real-time growth simulation and\\ resource optimization in rice farming.} & \makecell{Only focuses on simulation optimization\\without
extensive real-world validation}\\ \cline{2-7} 
    
    & \makecell{Smart Food Production\\and Supply} & \cite{maheshwari2023digital} & Computing server & \makecell{IIoT network} & \makecell{An ABS method with MILP fo\\optimizing food processing strategies,\\enhancing supply chain efficiency.} & \makecell{Fixed procurement and delivery\\processes has limit flexibility, and\\reliance on initial conditions requires\\larger datasets for better accuracy.} \\ \cline{2-7} 
    & \makecell{Smart Food Production\\and Supply} & \cite{karadeniz2019digital} & Computing server & \makecell{IIoT network} & \makecell{A eGastronomic things platform using\\IoT, AR, and VR technologies to\\enhanced secured funtions monitoring\\and operational simulation simulation\\for industrial ice cream machine} & \makecell{Limited validation beyond the\\specific ice cream machine and scope for\\broader gastronomic device applications.} \\ \cline{2-7} 
     \hline

\multirow{6}{*}{\rotatebox{90}{\makecell{ Space Industry}}}
    & \makecell{Spacecraft Industry} & \cite{yang2022digital} & Computing server & \makecell{Assembly node\\sensor network} & \makecell{A DT-driven system for\\spacecraft assembly, enabling real-time\\monitoring and simulation of assembly\\processes with virtual-real mapping.} & \makecell{The real-time system's effectiveness\\in complex assembly scenarios and its\\scalability to different spacecraft designs\\has not been validated.}\\ \cline{2-7} 
    & \makecell{Spacecraft Industry} & \cite{pinal2024mixed} & Computing server & \makecell{ISS propulsion and\\navigation systems,\\spacesuit sensors} & \makecell{A platform "DT of rice" for\\real-time growth simulation and\\ resource optimization in rice farming.} & \makecell{The integration's performance in\\diverse training scenarios and\\its scalability to other spacecraft\\systems remain partially explored}\\ \cline{2-7} 
    
    & \makecell{ Satellite Industry} & \cite{leutert2024ai} & \makecell{AI and ML-\\based servers} & \makecell{Industrial IIoT\\network} & \makecell{An in-orbit factory concept\\with digital twins, AI fault detection,\\and robotic assembly for\\autonomous satellite production.} & \makecell{AI-driven processes and\\real-time performance in\\space conditions are\\not fully validated.} \\ \cline{2-7} 
    & \makecell{ Satellite Industry} & \cite{zhao2022interlink} & Computing server & \makecell{Satellite-terrestrial\\ network} & \makecell{An INTERLINK with ASHER for\\optimized satellite handovers and\\a time-varying graph-based\\ITO for efficient routing, improving\\service continuity and data delivery.} & \makecell{The effectiveness of INTERLINK\\ in highly dynamic satellite\\environments and large constellations\\has not been extensively validated.} \\ \cline{2-7} 
     \hline

\multirow{6}{*}{\rotatebox{90}{\makecell{ Oil and Gas Industry}}}
    & \makecell{ Project Management\\\& Safety Assessment} & \cite{lai2022digital} & Cloud computing & \makecell{IIOT sensor\\network} & \makecell{A DT-based real-time\\monitoring, integration of\\AI for fault detection of\\supply chain strategies.} & \makecell{Effectiveness in managing new\\well types and extreme conditions\\has not fully tested.}\\ \cline{2-7} 
    & \makecell{Project Management\\\& Safety Assessment} & \cite{brewer2019digital} & Cloud computing & \makecell{IIOT sensor\\network} & \makecell{A FieldTwin scheme for\\ real-time data integration and\\ visualization, improving data\\accessibility and field management.} & \makecell{The platform’s performance\\with varied legacy data\\ sources and integration\\challenges remains underexplored.}\\ \cline{2-7} 
    
    & \makecell{Production\\Optimization} & \cite{shen2021digital} & \makecell{Neural network\\server} & \makecell{Oilfield sensor\\ network} & \makecell{A DT model for oil and gas\\ production, leveraging IIoT,\\real-time data, and AI to\\improve system efficiency by 3\%.} & \makecell{Limited exploration of DT\\integration across different phases\\of oilfield life cycle and\\varying production conditions.} \\ \cline{2-7} 
    & \makecell{Production\\Optimization} & \cite{latif2019holistic} & Computing server & \makecell{IIOT sensor\\network} & \makecell{An IAM-DT model to\\ extend gas field production\\ and optimize costs.} & \makecell{Adding new wells and compressor\\reconfigurations need more\\economic justification and\\involve operational risks.} \\ \cline{2-7} 
     \hline
\multirow{6}{*}{\rotatebox{90}{\makecell{ Robotics Industry}}}
     & \makecell{Industrial Robotics} & \cite{hoebert2019cloud} & Cloud computing & \makecell{Robot sensor\\network} & \makecell{A cloud-based DT with\\ontology for automated robot\\configuration, enhancing manufacturing\\flexibility and cost-efficiency.} & \makecell{The scalability of the system\\in diverse and complex manufacturing\\environments requires further validation.}\\
     \cline{2-7} 
    & \makecell{Industrial Robotics} & \cite{garg2021digital} & Unity & \makecell{FANUC robot\\(M-10iA/12)} & \makecell{A Unity-based DT platform\\for remote robot programming,\\ensuring low latency and\\high trajectory accuracy.} & \makecell{The model's accuracy and\\performance are only\\validated through simulations,\\ lacking real-world testing.}\\ \cline{2-7} 
    
    & \makecell{Autonomous Vehicle\\and Drones} & \cite{denk2022generating} & Computing server & \makecell{UAV and AV\\ sensor network} & \makecell{A homotopic shrinking\\based DT model using\\for robust path planning in\\dynamic 2D and 3D environments.} & \makecell{Effectiveness of the method\\in real-world scenarios\\and with varying dynamic\\conditions remains unvalidated.}\\
     \cline{2-7} 

    & \makecell{Autonomous Vehicle\\and Drones} & \cite{wang2024smart} & \makecell{Cloud computing} & \makecell{CAVs and RSUs\\sensor network} & \makecell{An SMDT platform with cloud\\services and a novel navigation\\system to improve autonomous\\driving and traffic efficiency.} & \makecell{The real-world performance\\ and long-term effectiveness of\\the SMDT platform in diverse traffic\\scenarios are still under evaluation.} \\ \cline{2-7} 
     \hline

\end{tabular}
\end{adjustbox}
\label{tab:dt_applications}
\end{table*}

\subsection{Robotics Industry}
Modern industries are becoming more and more dependent on robotics, which means that sophisticated instruments are needed to improve productivity, precision, and adaptability. With its virtual version of real robotic systems like actuators, robotic arms, grippers, and so on, DT technology presents itself as a game-changing solution. This technology enables precise management of robotic operations and predictive maintenance via real-time monitoring, modeling, and optimization. \cite{huang2021survey}. By modeling a robotic system digitally, companies may anticipate problems, maximize efficiency, and adjust to new situations without interfering with real-world operations. DT integration in robotics not only improves robot performance but also stimulates automation innovation, opening the door for more sophisticated, adaptable, and effective robotic systems in a variety of industries. Here, we focus on analyzing the roles of DT in industrial robotics, autonomous vehicles, and drones.
\begin{figure}[h]
     \centering
     \includegraphics[width=\linewidth]{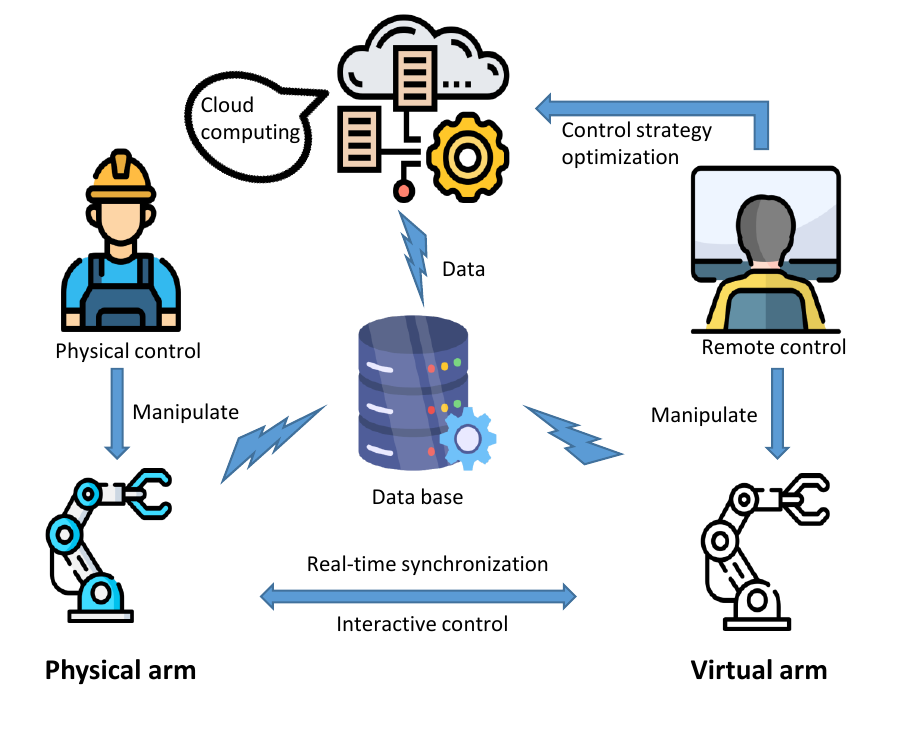}
     \caption{  A DT-driven framework for synchronized control of physical and virtual robotic arms via cloud computing.}
     \label{robotics}   
\end{figure}
\subsubsection{DT for Industrial Robotics}
The high costs of robot programming and reconfiguration, especially for small and medium-sized businesses (SMEs) provide quite a challenge. To reduce this, DTs offer real-time tracking and modeling for robots and their surroundings, lowering the related costs. The study \cite{hoebert2019cloud} presents the use of an ontology as a knowledge base to autonomously set up the DT and robot by defining their 3D environment. 
Robotic sensors wirelessly connect over cellular networks to share data with the cloud server. This approach's practical benefits are demonstrated by testing the suggested architecture in an industrial manufacturing setting, specifically for building THT devices on a PCB. 
Another study \cite{garg2021digital} develops a DT framework to improve robotic cell programming through online or remote interactions. It includes a Unity-based digital model and a FANUC (M-10iA/12) robot. Connectivity is established by linking robotics sensors to cloud computing via wireless communication, simplifying trajectory adjustments and reducing programming complexity. With a delay of about 40 milliseconds and minimal error in joint movements, the VR integration with the DT model guarantees consistent performance from both the digital and physical robots. Findings show that this DT approach works effectively for industrial applications and provides a more effective way to program and control robotic systems. Recently an exciting development of a digitally simulated Cyber-Physical and intelligent robotics laboratory that simulates 6-axis industrial robotic arm functionalities such as PTP, LIN, and CIRC motions is discussed in \cite{erdei2022design}, as shown in Fig. \ref{robotics}. Enabled by an integrated human-machine interface, this virtual lab allows users to build programs that incorporate these movements. SMEs can profit greatly from this system since it offers an inexpensive alternative to costly in-house training associated with installing, maintaining, and repairing actual robots. Its digital nature enables for extensive modification, with extra robot units that may be added or withdrawn, as well as pre-configured training or production areas based on unique facility requirements. Additionally, the virtual format eliminates room capacity restrictions, enabling numerous learners to participate at the same time. The program's teaching effectiveness was evaluated, and it was found that using DT machine units improves learning outcomes by providing a more interactive and efficient experience than traditional non-interactive demonstrations, even if the digital lab cannot fully simulate physical aspects such as space or weight. To enhance robotic control and simulation accuracy, the study \cite{xu2021digital} suggests another framework for DT-based Industrial Cloud Robotics (DTICR), which combines cloud computing and industrial robots (IRs). Physical robots, digital replicas, robotic control services, and DT data make up is the four main parts of this framework. The capabilities of robotic control are encapsulated in Robot Control as-a-Service (RCaaS), which is assessed and transferred to real robots via digital model simulations. By adding sensory data from the physical robots to the digital models, the system makes sure they are always up to date. The results demonstrate how efficiently DTICR synchronizes digital and physical robots, allowing for accurate control and situational adaptation. This approach demonstrates flexibility and extensibility through ontology models, making it a robust solution for advanced industrial robotic applications. Even more, the authors in \cite{liu2022digital} use Deep Reinforcement Learning (DRL) to improve the training of industrial robots for tasks like precision manipulation, with a focus on assembly-oriented grasping. Using a DT model, they deal with the challenge of transferring learned skills from virtual settings to real-world applications. This method entails developing a constantly updated digital model of the physical robotic system, which works alongside the actual robot. For guiding the robot's movements, visual data is processed by both the digital and physical systems. The robot's movements are adjusted based on the outputs from the digital model, resulting in more precise gripping. The results validate that the use of a DT with DRL greatly enhances the efficiency of robotic skill transfer from simulation to practical environments, ensuring increased accuracy in industrial applications.

\subsubsection{DT for Autonomous Vehicle and Drones}
DT technology is proven to be crucial for managing complex and dynamic situations in the domain of autonomous vehicles and drones. By acting as a dynamic digital equivalent of physical systems, DT can make it possible to simulate, monitor, and modify these systems' functions in real-time. This capability is especially vital for drones and autonomous vehicles, which have to navigate in constantly changing environments and complete several tasks at once. Traditional pathfinding algorithms frequently fail when confronted with real-world complexity, such as variable passage widths and the requirement for frequent recalibration. 
To overcome these limitations, the authors in \cite{denk2022generating} introduced a technique called homotopic shrinking to generate comprehensive DTs. This method creates a variety of paths, considering not just the shortest route but also the width of passages and other environmental restrictions. By connecting UAV and AV sensor networks to a computing server, the study demonstrated this strategy's success through simulations on 2D and 3D maps, showcasing its flexibility in handling diverse barriers and abrupt shifts.
Similarly, this study \cite{wang2024smart} proposes an innovative Smart Mobility DT (SMDT) architecture that improves the management of Connected and Autonomous Vehicles (CAVs) in next-generation wireless networks. The SMDT platform optimizes autonomous driving by leveraging cloud-based services and combining modern technologies such as CAVs, RSUs, and cellular V2X (C-V2X). The platform includes a new navigation system that enhances road safety and traffic efficiency by utilizing data from DTs. Through proof of concept experiments, the authors demonstrate that the SMDT system effectively reduces average travel times and minimizes delays caused by traffic incidents. Results show that the system meets 3GPP standards, with peak latencies of 155.15 ms for DT modeling and 810.59 ms for route planning, proving its compatibility with emerging V2X requirements.

Another important aspect of this sector is maintaining driving safety. In this work \cite{wang2022digital2}, a DT approach to highway driving safety analysis is presented. The approach consists of three primary components: gathering real-world vehicle data, modeling vehicle motions virtually, and evaluating driving risks. Using drone footage, vehicle movements are first captured, and then machine vision algorithms are used to extract exact trajectories. Afterward, DT simulates the dynamics of vehicles and roads in a virtual environment using these data points. Case study results showed that this approach evaluates numerous driving risks, including the possibility of sideslip, rollover, and crashes, based on the stability and deviation of vehicle trajectories.
Furthermore, the development of smart ports has become critical as ports throughout the world implement automation more and more to fulfill rising cargo needs and improve operational efficiency. To address port congestion issues and the necessity for precise navigation, the authors introduce \lq TwinPort\rq\, a DT model that uses drone-assisted data collecting within a 5G network framework in \cite{yigit2023twinport}. By combining real-time data from drones to direct ship navigation and maneuvering, TwinPort provides a sophisticated port management solution. A recommendation engine is built into the system to enhance the efficiency and accuracy of the trajectory by optimizing the docking procedures. Experimental results show that TwinPort improves navigation to follow the shortest path and decreases fuel consumption and operational expenses, thereby contributing to environmental sustainability.
\section{Security and Privacy Issues in Industrial DTs} \label{SectionV}
Although DT technology is emerging as a promising solution for industrial systems, there are still a series of challenges related to privacy and security. This section presents security issues on three degrees: physical, digital and communication, as well as human interfaces, along with countermeasures.

\subsection{Physical-Level Security Issues} 


\subsubsection{Operational Software Attacks}
The IIoT devices consist of their hardware, operating system, and operational software (called OT). DTs rely on the combination of a variety of OTs from different vendors. Hence, bugs in their code lead to a variety of potential security risks, including memory reverse-engineering attacks \cite{9864200}, buffer overflow exploitation, manipulation attacks, and node behavior-based attacks. 
In general, studies demonstrated that most OT devices are capable of being attacked in one of the above methods through potential vulnerabilities in their operating systems, especially Original operating systems without updated patches of manufacturers. Through combining industrial monitoring software into programmable machines and logic controllers (PLC) and rootkits for controllers of OTs, worms/malware easily penetrate IIoT systems, as well as launch attacks on part or all of DTs.

\subsubsection{Privilege Escalation}
In any industrial setting, when an attacker gains access to the OT domain, they may use different methods for privilege escalation towards taking over the administrator's rights to completely control the system through disconnecting 1-layer devices, changing configurations, generating incorrect values, controlling the system, and making extremely robust impacts to up layers. For instance, the Rowhammer is a strict hardware vulnerability in the DRAM memory of IIoT devices. By repeated accessing to hammer rows, the attack can induce bit flips to perform a privilege escalation attack for administrator privileges \cite{9919335}. In addition, this method can also provide false information for AI aggregators to generate incorrect input and output values of both physical and digital spaces in DTs.
\subsubsection{Rogue IIoT/CPS Devices}
In DTs, the in-network OTs can be accessed on their own for industrial deployment, cloning, and replacing devices. Attackers can install and execute malicious codes to take control of the physical space or the digital space as presented. The study in \cite{10367998} presents a rogue access point attack to collect information and privacy of system users. The work in \cite{10454724} also shows security vulnerabilities for performing industrial computer platform attacks based on rogue IIoT devices in the industry domain.
\subsubsection{Extracting System Information}
The in-networks users have complete rights to extract system information from their own IIoT/CPS for private information authentication or set up security parameters of DTs. By leveraging this information, attackers can analyze network traffic, determine the server IP address, exploit system parameters, or provide false information to attack both physical and digital spaces. The work in \cite{9389670} shows model extraction attacks in ML-based IIoT systems. By executing the same machine learning model on different IIoT devices, attackers can steal parameters and ML models to perform attacks in DTs.

\subsection{Digital- and Communication-Level Security Issues}
We here discuss security issues related to industrial computing and virtualization infrastructures, and communication.





\subsubsection{Privilege Escalation}
Through security vulnerabilities, the attackers perform escalating privilege attacks within the virtualization system. Then, they can navigate data threads and virtual resources and initial multiple attacks to exfiltration, manipulations, overflows, or analysis of databases on cloud servers. Similarly, escalated privilege attacks through VMs and network nodes infected with malware may also attack other legitimate virtual resources of the system. For instance, in \cite{RUBIO2019101561}, the authors designed an escalated privilege attack by exploiting the virtual channels with connection to the virtual network inside the VMs.

\subsubsection{Rogue Virtual Resources}
In this method, the attackers perform privilege escalation attacks to access the virtual industrial servers, then insert, clone, or replace legitimate resources with malicious resources for the main purpose of taking control of a part or the entire DT model consisting of both physical and digital spaces. The study in \cite{9170997} shows a rogue virtual resources-based attack on edge computing servers. Furthermore, the work in \cite{9048618} presents derived attack methods from rogue virtual machines to legitimate virtual resources, virtual IP/MAC spoofing, or manipulation data threads.


\subsubsection{Virtual Resource Tampering}
Through privilege escalation attacks, attackers can manipulate the services of the DTs, even change the synchronization process of the digital models, modify the behavior of both physical and digital spaces, and attack edge servers. Attackers can also control virtual machines that store DT logic, manipulate digital resources and the hypervisor, create channels to steal intellectual resources, and inject malicious codes for in-depth attacks \cite{9020077}. Obviously, the attack method has a serious impact on the seamless DT services and the privacy of data and end-users.



\subsubsection{Privacy Leakage}
By this method, attackers can steal sensitive data from industries, or organizations such as production, logistics, marketing plans, and customer databases \cite{10409579}, \cite{10239369}. In addition to data privacy, organizations face different risks, such as operating models and system configuration information \cite{10264857}. Moreover, location privacy also should be considered. The location of the cloud or edge server that contains hypervisors and DT's logic can be the first target for attackers \cite{9628062}. Furthermore, depending on how the architecture and infrastructure of DTs are connected, the resource sharing and allocation schemes, and the management hierarchy, attackers can monitor to perform the next privilege escalation attacks.

\subsection{Human-Machine Interfaces (HMIs)-Level Security Issues}

\subsubsection{OT Attacks}
HMIs consist of OT components such as OS, communication interfaces, and human-machine interfaces to control, manage, and show results. It allows humans to interact with the physical space of DTs. Consequently, these characteristics make them face several vulnerability risks, as presented in \cite{10234712}, \cite{9954177} including (i) changing the accuracy, trust, and integrity of the represented object; (ii) stealing production data and customer databases; (iii) change the direction of data flows to steal system data; and (iv) install malicious codes and resources to prepare for next privilege escalation attacks.

\subsubsection{Rogue HMIs}
It is noted that end-users have legal rights to access full OT domains. In this way, attackers as end users may insert malicious code, misconfigure or replace OT devices, and clone HMIs with only a connection to the DTs. Through these rogue HMIs, attackers may perform typical attacks. (i) provide invalid/misinformation values for inputs/outputs of DTs; (ii) change data values of representation objects in the HMI; (iii) Extract system information; (iv) Navigate data flows to extract/steal factory production data; (v) Disable or delay maintenance of HMIs. In the research, \cite{8737455} designs a rogue WIFI interface that has SSID, MAC, and IP address identical to the legitimate devices. The authors demonstrated that this attack method may insert malicious codes, generate fake messages to manipulate or steal data, and is difficult to countermeasure, relying on existing security mechanisms.
\subsubsection{Visualization Tampering}
According to this method, attackers may modify specific HMI settings and services for final visualization tampering of representation objects. In this way, attackers may hide information, show misinformation, or change the data integrity of digital objects. For instance, In the study \cite{10607909} presents a tampered image attack. This causes the HMI to present a different reality and forces the system makes an incorret decision.


\subsection{Countermeasures}
\begin{figure}[!b]
    \centering
    \includegraphics[width = 0.45\textwidth]{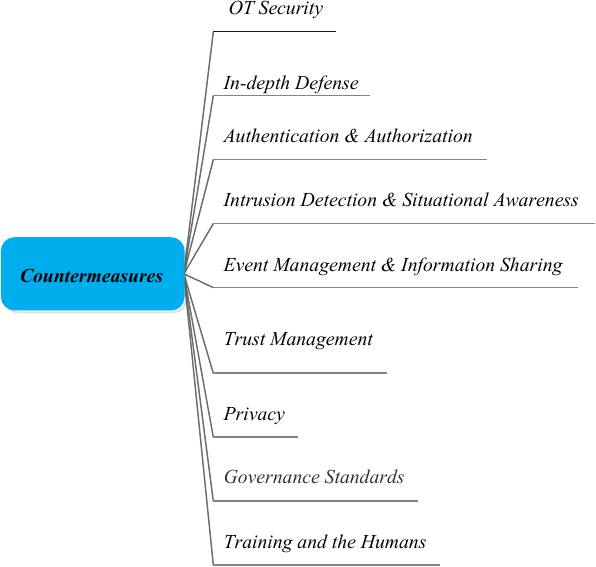}
    \caption{Classification of countermeasures to DTs.}
    \label{Sec3-Quy}
\end{figure}

In the previous section, the potential of DTs in the industry domain is demonstrated. However, there are still a series of concerns related to privacy and security in order to realize pervasive DTs. To address this problem,  countermeasures (summarized in Fig.\ref{Sec3-Quy}) are needed to ensure reliable and safe industrial DT systems.

\subsubsection{OT Security}
The operating DT relies on combining OTs that consist of IIoT devices, its operating system, and software. It's unfortunate that these 1-layer elements may contain endogenous vulnerabilities. This may be due to a missing design or inadequate validation, especially in the case of third-party resources. In our vision, the DT across industries should be designed to (i) use a trusted platform module, combine a high-reliability deployed environment, (ii) use secure source codes, (iii) high standard security framework, and (iv) strict authenticate process for OT devices or servers). Accordingly, the work in \cite{10090432} presents recent advances related to OT security in aspects of architecture and solutions for DTs. Similarly, in \cite{10496463}, the authors mention privacy and security for IT/OT devices in 6G networks.

\subsubsection{In-depth Defense}
This approach focuses on protecting machines, physical objects, servers, and virtualization systems. In this case, in-depth defense forms a basic platform for security DTs and vice versa. Hence, security solutions must be combined to protect DTs. In our vision, this solution has a role as first protection, therefore, function segmentation and separation are feasible approaches. This is based on the exploitation of in-depth network technologies such as firewalls, VLANs, virtual private networks, IDSs/IPSs, and good practices. On the other hand, to counter the interference of insiders; frequent monitoring operations are needed, including monitoring the commands to the servers, the hypervisor memory management, and security of the virtual machines and their functions. In the study \cite{10239369} presents a multi-layer protection framework, including physical, cloud, and edge environments for 6G-based DT in Industry 5.0.

\subsubsection{Authentication and Authorization}
The combination of real-world objects and digital space makes DTs complex. This requires (i) real-time authentication related to identifying users and data and (ii) authorization management in both physical and digital spaces \cite{9852383}.
It's noted that the integration of authentication and access policies forms the first security perimeter of DTs. This may be executed on OT devices or edge/cloud servers. Authentication requires entities to verify their legal access to system resources from both physical and digital spaces \cite{10522584}. Moreover, powerful authorization mechanisms are necessary depending on the scale and deployed context of DTs. For instance, the work in \cite{9430900} introduces a context awareness mechanism to detect security issues and auto-update authorization policies based on ML techniques and blockchain. In our vision, if access rights to these resources are unguaranteed, attackers can perform escalated privilege attacks. Furthermore, authorization policies are vital for DTs. In reality, several results based on these approaches have demonstrated their feasibility for scale-large IIoT systems, and they may be adjusted more suitable to DTs.

\subsubsection{Intrusion Detection and  Situational Awareness}
Thanks to ML algorithms, DTs can perform industrial data analysis for situational awareness and detect early potential security risks.  DTs may detect what is happening ubiquitous with a high accuracy rate, they can detect the threat with details such as location and impact degree and provide the real-time origin traceability of attacks. However, studies in \cite{9302783}, \cite{10000804} demonstrate that anomaly detection and attack traceability need to process a vast amount of data. Hence, it still poses significant study challenges. From the situational awareness perspective, through data collection, visualization principles, and consensus-based mechanisms, DTs may make situational awareness and decision-making correct \cite{9016083}. In our vision, the complexity of DTs forms from the diversity of data sources, operating environments, and devices. This leads to diffirent anomalous events. On the other hand, the DTs are usually combined with third-party vendor systems, more increasing the security risks. Thus, administrators should be aware of the full system operation.

\subsubsection{Event Management and Information Sharing}
One of the leading security countermeasures of DTs is through security operations centers (SOCs) that are monitored by professional administrators. SOCs use security information and event management systems (SIEMs) to form a full picture of related security issues and discover vulnerabilities of DTs. In this way, any suspicious action in a DT may be considered. The research in \cite{9864249} demonstrates a consistency-checking method of DTs based on observed timed events in the manufacturing domain. For information sharing, distributed ledger technology can be a suitable option for tracing the actions taken by objects in digital space. This guarantees high availability and transparency. The study in \cite{10272999} presents a collaborative computing scheme based on blockchain for DTs in the vehicular domain.

\subsubsection{Trust Management}
Forming a reliable operating industrial environment is one of the countermeasures to DTs' security issues. By establishing trust between objects, components of a DT can securely interact and exchange industrial data, ensuring the reliability, integrity, and confidentiality of the digital-physical system. The work in \cite{10488084} proposes an enhanced reputation scheme to estimate the trust of objects and industrial components in the DT. The component's behavior and events are evaluated by a reward or penalty mechanism. The score deviation makes changes related to the trust of a component in DT. However, this method also increases overload due to requiring a high level of computation and storage since they usually need to use past conducts and reflect them in the current trust. In this process, these solutions can also demand a significant exchange of information between components to compute trust levels. In our vision, over these inconveniences, the integration of trust mechanisms is necessary to improve decision-making and to detect anomalous behavior within DT.

\subsubsection{Privacy}
Aiming to realize pervasive DTs, privacy is one of the vital factors for DTs. Privacy can be leaked in ways such as IP addresses of DT servers, locations, and data. DTs need to be aware of what data may be shared with different degrees to access the information between DT's components and objects. In the industry domain, the policy uses DT's resources, which requires us to consider this method under other aspects. Due to operational processes IIoT/CPS devices generally perform the same operations according to routine movements and actions. This allows attackers to infer resources, behaviors, and locations related to privacy data. The study in \cite{10269659} shows the need for location privacy and route protection together with anonymity approaches to protect the identity of the physical devices and virtual resources.

\subsubsection{Governance Standards}
In the industrial domain, organizations need to consider DT security issues, including safeguarding physical resources, LAN, devices, access policies, edge and cloud computing infrastructure, issues related to humans under governance standards, and related legal procedures. This also should be considered throughout the DT life-cycle. Regarding this, the ISO organization has developed and announced the set of ISO 23247, parts 1-4 standards for the implementation of DT in the industrial domain, the set of ISO/ IEEE 11073 standards for Healthcare in smart cities \cite{9108291}, etc. In our vision, these standards improved governance and security DTs. However, there are still many related security issues that need to be addressed due to the complexity of deployment contexts in the industrial domain.

\subsubsection{Training and the Humans}
DT is an emerging technology thanks to the breakthrough development of communication technologies, the Internet of Things, and advanced AI techniques. From this perspective, end-users and administrators should be continuously updated and enhance their knowledge to ensure the deployment and exploitation of DTs effectively and sustainably. One of the effective training solutions is through regular training courses under personalized and integrated educational methodologies. Like with general security systems, humans are the heart of DTs. In another aspect, they are also the most dangerous attackers, especially insiders. This is beyond the control of security systems. In our vision, frequent HR assessment strategies should be used to detect insider, intentional actions, or human errors. The study in \cite{10.1145/3303771} presents a comprehensive review of human aspects that play an insider threat to DTs and detailed countermeasures.

\section{Key Research Findings and Future  Directions} \label{SectionVI}

 \subsection{Key Findings} 


 \subsubsection{DT Services in Industries}
  DT services are quickly becoming essential for modernizing industrial operations, providing tools for real-time monitoring, process optimization, and predictive maintenance. DT technology enables companies to manage operations more dynamically and effectively by converting physical assets and processes into digital versions. For instance, DT services in manufacturing allow for continuous line monitoring, promptly detecting and resolving problems to save downtime and increase efficiency \cite{aivaliotis2019use}. DT services in the energy sector offer comprehensive energy system simulations that aid in resource management optimization and make it easier to integrate renewable energy sources. Numerous studies on DT have highlighted that its adoption enhances industrial management by improving operational efficiency, enabling advanced predictive maintenance, facilitating better resource allocation, and increasing the accuracy of performance assessments. DTs provide industries with the information and understanding required to foresee issues and make wise decisions by providing a comprehensive, real-time perspective of operations. This capacity is particularly important in industries where even minor inefficiencies can result in significant losses. Furthermore, DT services guarantee that all the components of an industrial activity are in perfect sync and function as a team to accomplish shared objectives. By combining machine learning and advanced analytics, DTs can also predict future trends, enabling businesses to solve problems proactively and continuously improve their procedures \cite{muller2020dynamic}. 

\subsubsection{DT Applications in Industries}
The application of DT technology offers a range of advanced capabilities that considerably enhance industrial operations. Upon reviewing the current literature, we find that DT applications help companies by solving several important problems, including resource management, predictive maintenance, and operational efficiency. Centralized processes are often the foundation of conventional industrial systems, which can lead to inefficiencies and delays in problem-solving. DT technology solves these issues with a dynamic, real-time virtual model of physical assets and processes.  To enhance resource management and integrate industrial sources, DTs are utilized for grid management and optimization \cite{baboli2020measurement}. More efficient energy distribution and responsiveness to variations in energy consumption are made possible by this feature. Similarly, optimizing energy storage and integrating renewable sources into power systems \cite{soderang2022development},  thermal energy storage systems in smart buildings \cite{lv2023digital}, smart water management Platform \cite{alves2019digital}, eGastronomic things \cite{karadeniz2019digital}, digital astronaut training \cite{pinal2024mixed}, FieldTwin in oil industry \cite{brewer2019digital} and many more. DTs also help with advanced predictive maintenance, which reduces downtime and lengthens asset lifespan by using real-time data from digital models to foresee equipment breakdowns and schedule timely maintenance.  
\subsubsection{Security and Privacy in Industrial DT}
DT technology is becoming more and more integrated into industrial processes, making strong security and privacy protocols vital. DTs include creating detailed digital duplicates of real assets and processes, which requires handling vast amounts of sensitive data \cite{hui2022digital}. If not properly secured, this data might become a major target for cyberattacks, resulting in potential breaches, illegal access, or modification of the digital models \cite{wang2023survey}. The transition from old centralized systems to DT-driven frameworks poses new risks, where a single point of failure may threaten a whole operation \cite{dietz2019distributed}. Addressing these security concerns necessitates a multilayered approach. To safeguard data integrity and make sure that only authorized users may access or change the DT, industries are starting to implement cutting-edge encryption techniques. Furthermore, research is being done to investigate how blockchain technology may be used to improve data transaction security in DT systems \cite{huang2020blockchain}. Blockchain's decentralized ledger provides a transparent and tamper-proof record of all interactions with the DT, facilitating change tracing and verification \cite{putz2021ethertwin}.
Protecting privacy is just as important as securing data, especially when it comes to confidential or private data. To keep unwanted parties from obtaining confidential data, businesses are using strict access control procedures and data anonymization techniques more and more. This is especially crucial for economy sectors where major financial losses or harm to one's reputation could result from data breaches. Furthermore, security management becomes much more complex due to the real-time functioning of DTs. It is critical to always make sure that these systems are secured against unauthorized access. Industries are implementing advanced firewalls and multi-factor authentication (MFA) to fortify the perimeters surrounding their DT environments \cite{10522584}. Using AI and ML to proactively identify and address possible risks is becoming more popular than only relying on these conventional security methods. Security systems powered by AI can continuously monitor DT operations, spot irregularities, and take quick action to reduce risks before they become serious cyber events. Additionally, integrating SOCs and utilizing SIEMs provide an in-depth knowledge of security concerns and vulnerabilities in DTs. Distributed ledger technology can help trace activities made by objects in the digital space, assuring high availability and transparency \cite{9864249},\cite{10272999}. Industry stakeholders need to be aware of what data may be shared and how much, as privacy considerations are very important. Trust management mechanisms, including reputation schemes and continuous HR assessments, help detect insider threats and ensure reliable operation \cite{10488084},\cite{10.1145/3303771}. As DT technology develops and becomes more deeply integrated into industrial processes, it is imperative to take a proactive approach to security and privacy to guarantee that the benefits of DT are fully realized without sacrificing safety.

\subsection{Future Research Directions}

\subsubsection{DT in 6G Networks}
DT technology is poised to greatly improve the capabilities of 6G networks, providing a new level of real-time monitoring and administration. With 6G expanding the limits of data transfer and communication, DTs offer an essential instrument for generating digital replicas of real-world systems, facilitating more accurate and dynamic network functions \cite{ahmadi2021networked}. As mobile communication services for the 6G Internet of Things are developed, the idea of the DT is becoming more and more important as an enabling technology \cite{masaracchia2022digital}.  Effective resource efficiency is crucial for 6G networks to use twin objects across a range of IoE applications. Allocating computing and communication resources intentionally is essential to maintaining the functionality of one twin-based IoE service without adversely affecting others. While allocating resources, like edge servers, to particular twin-based services is one approach, it may result in underutilization and inefficiencies. For all twin-based services to operate at their best, resource allocation must be balanced. Keeping mobile consumers' service experiences uninterrupted is also critical in a DT-powered 6G system. However, users' devices may disconnect from the base station or access point connected to their twin object when they relocate, which could cause service interruptions. Keeping the user connected to their twin object through a backhaul link is one way to address this problem, albeit there may be some small delays and disruptions. A more efficient method would be to apply machine learning techniques to improve the process of dynamically migrating services to new twin objects depending on the user's anticipated moves. As a result, even when users move between different coverage zones, the service will operate more smoothly and responsively. Since DTs operate in real-time, it is possible to make quick modifications and improvements, which is essential for handling the complexity of 6G situations. For instance, mobile crowdsourcing-enabled 6G mobile network video streaming with DT \cite{qi2023digital}, DT-powered, tiered security architecture for IIoT settings enabled in 6G \cite{alcaraz2023protecting}. Furthermore, new opportunities for spectrum management are created by the integration of DT technology with 6G networks like blockchain and low-latency federated learning for edge association in 6G networks powered by DT \cite{lu2020low}. Real-time simulations can guarantee the secure and effective allocation of spectrum resources. The incorporation of DTs can offer a continuously updated network perspective, which aids in the prompt identification and response to security risks, hence improving the security posture of 6G networks.  Other significant works also propose the flexible edge connectivity for wireless DT networks in 6G. Network infrastructures will become more intelligent, flexible, and resilient as a result of the integration of DT technology with 6G.

\subsubsection{DT with Big Data}
Major breakthroughs in industrial operations are expected to come from the combination of big data and DT technologies. With the continued production of vast volumes of data by various industries, DTs provide a means of utilizing this data to generate accurate, real-time digital copies of physical systems. Industries may delve deeper into complicated data sets by integrating DT with big data, which enables more precise predictive maintenance, astute resource management, and enhanced decision-making abilities. The development of scalable algorithms and systems capable of managing and analyzing the massive volumes of data produced by industrial processes is one of the main research directions. This involves increasing the speed and accuracy of data processing so that DTs can quickly deliver insights and adapt to real-world changes. Furthermore, achieving success in resolving issues with data security, integration, and quality is essential to reaping the full benefits of DTs when utilized with big bata.   Future studies should look into ways to provide seamless interoperability between DT platforms and big data systems, allowing for smooth data flow and integration across multiple industrial sectors. This will probably entail working to establish open-source tools that can be widely used, enhance data governance, and standardize procedures. big data and DT together will advance these fields and create more intelligent, responsive, and efficient industrial systems that will drive innovation and competition in Industry 4.0 and beyond.

Furthermore, converging DT technology and big data are becoming more and more important as companies shift toward more digitized operations. DTs can be crucial to the efficient analysis and use of the massive amounts of data generated by IIoT devices in smart cities. Through the integration of big data analysis with deep learning techniques, cities may optimize infrastructure, increase urban life quality, and improve resource management. Using big data analysis, DTs can handle large datasets quickly and accurately, giving them a dynamic and detailed picture of urban surroundings. For instance, in smart cities, the monitoring of different city services, such as energy distribution and traffic control, is made possible by the integration of IIoT-generated data with DTs. The ability of deep learning models to identify patterns and predict future trends aids in the decision-making of municipal planners. Additionally, the integration of DTs and big data makes it easier to develop prediction models that can replicate various events, including variations in traffic patterns or spikes in energy consumption. This capacity for prediction is essential for proactive resource management in cities and for reducing possible problems before they become more serious. Enhancing DT systems' scalability to manage the increasing volume of data from IIoT devices should be the key goal of future research in this field in order to maintain the responsiveness and efficiency of these systems. By expanding the integration of DTs with big data, particularly in smart cities, we can open up new opportunities for urban management, making cities smarter, more sustainable, and better prepared to face future challenges.

\subsubsection{Standardization for DT}
Standardization is becoming a key concern for the application and growth of DT technologies, especially in the manufacturing industry. The requirement for consistent standards becomes apparent when DTs are incorporated more deeply into production processes. The lack of uniform frameworks and protocols may jeopardize the compatibility of different DT systems and applications, resulting in inefficiencies and heightened complexity. Standardized DT architectures could be revolutionary in manufacturing, where efficiency and precision are critical. They would enable unified data exchange and process optimization by facilitating smooth connection across various systems and devices. This could improve a number of manufacturing-related factors, including resource management, predictive maintenance, and overall production efficiency. Moreover, developing common DT procedures could accelerate the technology's adoption throughout the company's operations. More concise integration pathways would help manufacturers because standardization of methods would guarantee that new systems work with both current infrastructure and emerging technologies. This could promote innovation and preserve a competitive advantage by making DT solutions more dependable and scalable. Next-generation DT designs should focus on creating thorough, industry-wide standards. To enable the efficient deployment and utilization of DT technologies throughout the industry, these standards must be flexible enough to accommodate the various demands of various manufacturing settings. 


\subsubsection{DT with Quantum Computing} \label{SectionVII}

The combination of DT technology and quantum computing is poised to unleash significant advances in a variety of fields, including network management and cybersecurity. As quantum computing advances, merging it with DTs could provide game-changing improvements in data analysis and system capabilities. One rising area of study is DTQFL (DT-assisted quantum federated learning). To enhance intelligent diagnostics in 5G mobile networks, this innovative method combines the advantages of DTs and federated learning \cite{rahman2024improved,nguyen2021federated} in the era of quantum computing. Through DTQFL, network optimization, and predictive maintenance will be improved by utilizing quantum computing's superior capacity to handle and process large datasets \cite{qu2023dtqfl}. Quantum computing and DTs work together to simulate real-time network conditions, which could result in 5G systems that are more resilient and adaptable and improve overall network performance and dependability.  DARIUS (DT-Assisted Robust Quantum Key Distribution) is another modern breakthrough. This program aims to promote quantum key distribution (QKD) by utilizing DTs to improve the security and performance of quantum communication systems. To establish secure communication, QKD relies on quantum mechanics; DARIUS seeks to strengthen and optimize QKD procedures by integrating DTs. DTs provide a virtual representation of the quantum communication ecosystem, allowing for better control and fine-tuning of key distribution processes. This integration has the potential to dramatically increase the robustness and efficiency of quantum cryptography networks. Future research should focus on how DTs and quantum computing may solve complex issues and provide new opportunities. Exploring advanced principles such as DTQFL and practical applications like DARIUS could lead to significant advances in network intelligence and secure communication. As the field advances, these novel combinations of DT and quantum technologies have the potential to significantly improve system capabilities and security in an increasingly digitized environment. 
\section{Conclusion}
\label{sec:conclude}
DT is an emerging innovation that has sparked significant interest for its ability to revolutionize industrial operations and enhance efficiency through the seamless integration of physical and digital systems. In this article, we have investigated the potential of DT to facilitate smart industries through a state-of-the-art survey and detailed analysis of recent research in this domain. This work addresses the absence of a comprehensive survey on the services and applications of DT across industries. In order to bridge this gap, we have first introduced the recent advances in DT and its enabling technologies and provided insights into their integration in realizing smart industries. We have then provided an updated survey on the services of DT in a wide range of sectors, namely data sharing, data offloading and caching, integrated sensing and communication, resource allocation, wireless networking, and metaverse, with a particular focus on the nature of communication and networking protocols among systems, machines, processes and their digital counterparts. Subsequently, we have focused on discussing the latest developments in integrated DT applications across significant use-case domains, including healthcare, manufacturing, transportation, energy, agriculture, space, robotics, as well as oil and gas, highlighting the crucial role of communication and networking technologies in enhancing DT effectiveness. From our comprehensive survey, we have summarized and analyzed several key findings. Additionally, we have presented potential directions for future exploration. We believe that this article will foster greater interest in the field of DTs and inspire further research efforts toward fully realizing the potential of DT technology in transforming industrial practices.

\bibliographystyle{IEEEtran}
\bibliography{references}

\end{document}